%% file: main.tex
\documentclass{article}

\PassOptionsToPackage{numbers, compress}{natbib}

 \usepackage[eandd, final]{neurips_2026}

\usepackage[utf8]{inputenc} %
\usepackage[T1]{fontenc}    %
\usepackage{hyperref}       %
\usepackage{url}            %
\usepackage{booktabs}       %
\usepackage{amsfonts}       %
\usepackage{amsmath}
\usepackage{nicefrac}       %
\usepackage{microtype}      %
\usepackage{xcolor}         %
\usepackage{xspace}
\usepackage{graphicx}
\usepackage{caption}
\usepackage{paralist}

\newcommand{\datasetname}{ZeoBench\xspace}

\title{\datasetname: A Benchmark for Self-Supervised Learning on 3D Zeolite Representations}

\author{%
  Aaron Sun \quad Yachan Liu \quad Ping Yang \quad Peng Bai \quad Subhransu Maji\\
  Manning College of Information and Computer Sciences\\
  University of Massachusetts Amherst, MA 01003, USA \\
  \texttt{\{aaronsun,yachanliu,pinyang,pengbai,smaji\}@umass.edu} \\
}

\begin{document}

\maketitle

\begin{abstract}
Zeolites are an important class of porous crystalline materials with broad applications in gas storage, separations, and catalysis.
Although hundreds of thousands of experimentally realized and hypothetical zeolite structures are known, identifying materials optimized for a target application remains challenging because property labels are expensive to obtain and available only for limited structure--molecule pairs.
While prior work has demonstrated that 3D neural representations can predict zeolite properties effectively, it remains unclear which representation-learning strategies are most effective in the low-data regime relevant to materials discovery. 
We present \datasetname, a benchmark for self-supervised learning on dense 3D representations of all-silica zeolites, and evaluate the resulting features on six existing downstream adsorption-property prediction tasks, across a range of training set sizes.
Our study compares self-supervised 3D pretraining on volumetric zeolite data, adaptations of image-pretrained 2D ConvNets and vision transformers to 3D via multiview and channel-inflation strategies, and 3D ConvNets pretrained on out-of-domain volumetric datasets, alongside hand-crafted descriptors and supervised baselines. 
Using our pretraining and evaluation framework, we find self-supervised 3D representations outperform existing methods and achieve state-of-the-art performance across both in-domain label regimes and for the unseen domain of metal organic frameworks. 
Learned representations consistently outperform hand-crafted descriptors, and convolutional architectures are generally more effective than vision transformers in this setting. 
These results establish \datasetname as a practical benchmark for 3D representation learning in zeolites and provide guidance for data-limited adsorption property prediction. 
\end{abstract}

\vspace{-12pt}
\section{Introduction}
\vspace{-6pt}
\input{figs_tabs/fig_1}

Zeolites are nanoporous crystalline structures with wide uses in chemical industrial applications, including catalysis and adsorption.
In general, zeolites are periodic framework materials formed from corner-sharing tetrahedral building blocks, and their composition can vary depending on the framework atoms and extra-framework species. 
In this work, we focus on all-silica zeolites, whose frameworks are composed of repeating SiO\textsubscript{4} tetrahedra arranged into different pore and cage topologies (see Figure~\ref{fig:fig1}).
This fixed chemical composition makes them a natural setting for studying geometric representation learning, as variation in adsorption behavior is driven completely by differences in 3D pore geometry, framework topology, and the shape compatibility between the zeolite and the target molecule~\cite{smit2008molecular}. 
Although over 300,000 hypothetical zeolite structures have been proposed, only about 260 distinct framework topologies have been experimentally synthesized~\cite{IZSAC, pophale2011database}. 
An active area of research is determining zeolites with desirable properties for future synthesis~\cite{bai2015discovery, liu2026assessment}.

The interactions of these structures with target molecules can be accurately predicted through simulations, but at considerable computational cost for each pair evaluated~\cite{bai2013trappe, bai2015discovery, bai2014understanding}.
Existing works have found success in estimating these predictions through supervised learning with significantly faster evaluation~\cite{liu2023zeonet}.
However, these models are typically task-specific: a separate model is trained for each target molecule, and for some tasks many structures may be unsuitable or unavailable for supervised training. As a result, prior supervised studies use only a task-specific labeled subset of the available zeolites, leaving much of the unlabeled corpus unused~\cite{liu2025representation}.

Self-supervised learning (SSL) offers a natural way to exploit this unlabeled data for transferable, task-agnostic representation learning. 
SSL methods define pretext objectives on unlabeled examples, such as matching representations of different augmented views of the same input or reconstructing masked or corrupted inputs~\cite{he2020momentum, chen2020simple, caron2021emerging, grill2020bootstrap, he2022masked}. 
These methods have been highly successful in vision and language, where large-scale pretraining produces task-agnostic backbones that transfer effectively to downstream tasks~\cite{devlin2019bert, yang2025qwen3}. 
In the zeolite setting, SSL could enable models to learn reusable 3D structural features from the full corpus of unlabeled frameworks before adapting to specific tasks.

Although SSL has been studied for molecules and materials, most prior work has focused on graph, coordinate, or string-based representations~\cite{liu2021pre, zhou2023uni, sun20243d, chithrananda2020chemberta, ross2022large}. 
In contrast, all-silica zeolites provide a particularly clean testbed for dense 3D representation learning, because the structures share the same framework chemistry and differ only in geometry and topology. 
Recent supervised work has shown that dense 3D grid representations are effective for predicting adsorption in this setting~\cite{liu2023zeonet}. 
However, it remains unclear which self-supervised objectives, augmentations, architectures, and pretraining sources are most effective for dense 3D zeolite grids, especially in the low-label regimes most relevant to materials discovery.

To bridge this gap, we present \datasetname, a benchmark for SSL on zeolites.
We define standardized pretraining and evaluation protocol on six existing downstream adsorption-property prediction tasks and perform a detailed analysis on four popular SSL methods.
We make the following contributions:
\begin{compactenum}
    \item We introduce \datasetname, a benchmark for SSL on dense 3D grids of all-silica zeolites, built from 331k real and hypothetical zeolite structures, with fixed train/validation/test splits.
    \item We standardize six existing downstream adsorption-property prediction tasks with label-budget splits, evaluation metrics, and linear-probe and fine-tuning protocols.
    \item We benchmark four SSL objectives, out-of-domain 3D volumetric pretraining, ImageNet-pretrained 2D backbones adapted to 3D, hand-crafted descriptors, and supervised baselines.
    \item We show that in-domain SSL substantially improves low-label adsorption prediction, with contrastive methods producing the strongest frozen representations and SSL initialization improving fine-tuning performance.
    \item We conduct transfer learning experiments on metal organic frameworks and find \datasetname pretraining improvements generalize to the out-of-domain task and outperform existing methods and training from scratch. 
    \item We analyze augmentation and architecture choices, finding that scale augmentation is critical for SSL and that convolutional neural networks (CNNs) currently outperform vision transformers (ViTs) in this setting.
\end{compactenum}
The benchmarking code, task-specific training splits, and pretraining configurations are available at \url{https://github.com/cvl-umass/ZeoBench}.

\section{Related Work}

\subsection{Self-supervised Representation Learning}
Self-supervised learning (SSL) has become a standard approach for learning visual representations from large unlabeled datasets. We group the methods into three broad families: contrastive, distillation-based, and reconstruction-based methods. Contrastive methods, such as SimCLR~\cite{chen2020simple} and MoCo~\cite{he2020momentum, chen2020improved, chen2021empirical}, learn by bringing together embeddings of augmented views of the same input while separating embeddings of different inputs. Distillation-based methods, such as BYOL~\cite{grill2020bootstrap} and DINO~\cite{caron2021emerging, oquab2023dinov2, simeoni2025dinov3}, avoid explicit negatives and instead train a student network to match the representation of a momentum-updated teacher. Reconstruction-based methods, including masked autoencoding objectives~\cite{he2022masked, zhou2021ibot, assran2023self}, learn by reconstructing masked or corrupted inputs and are often studied with transformer-based architectures.

In this work, we focus primarily on contrastive and distillation-based methods. These methods are broadly compatible with CNNs, which we find to be more effective than vision transformers for dense 3D zeolite representations. Furthermore, reconstruction objectives do not always produce representations that are optimal for linear probing or low-label transfer; e.g., both~\cite{marks2025closer,he2020momentum} report substantially lower performance with linear probing than with end-to-end fine-tuning.

\subsection{Supervised Learning for Porous Materials and Zeolites}

Machine learning has been widely used for molecular and materials property prediction, including quantitative structure--activity relationship modeling in drug discovery~\cite{stokes2020deep, jin2021deep, vamathevan2019applications}, force-field and interatomic potential prediction~\cite{schutt2017schnet, batzner20223, behler2016perspective, Batatia2022mace}, and adsorption-property prediction in porous materials~\cite{gaillac2020speeding, anderson2020adsorption}. These applications illustrate the promise of learned representations for connecting atomic structure to target physical or chemical properties.

For porous materials, prior work has explored a range of structural representations. Descriptor-based approaches use hand-crafted features such as density, pore diameter, pore volume, surface area, bond lengths, and bond angles~\cite{gaillac2020speeding, anderson2020adsorption, liu2023zeonet}. Graph neural networks encode atoms, bonds, chemical connectivity, and interatomic distances~\cite{chen2022universal, xie_crystal_2018, schutt2017schnet}. Coordinate-based and point-cloud models operate directly on atomic positions~\cite{wang2019dynamic,liu2023zeonet}. Dense 3D volumetric representations, including energy grids, atomic grids, and distance grids, provide an alternative view that represents the geometry and chemistry of the material on a regular lattice~\cite{cho_nanoporous_2021,hung_chemistry-encoded_2022,bucior2019energy,li2021machine}.

Zeolites provide a particularly important testbed for studying these representations. They are crystalline microporous materials with broad applications in adsorption, separation, and catalysis, but their functional properties depend sensitively on pore geometry and chemical structure. 
Recent work has shown that dense 3D grid representations are highly effective for zeolite property prediction, including target-molecule adsorption and synthetic feasibility~\cite{liu2023zeonet,liu2025representation,liu2026assessment}. These results motivate our focus on 3D zeolite grids as the input representation for representation learning.

However, supervised learning for zeolite property prediction is limited by the cost of generating labels. 
Adsorption labels often require computationally expensive Monte Carlo simulations, especially for complex molecules, and many structures can also be inapplicable for supervised learning depending on the task~\cite{bai2012multicomponent,bai2015discovery}. 
As a result, prior supervised studies typically train on labeled subsets of zeolites, while a much larger set of unlabeled structures remains unused. 
Moreover, effective training often requires on the order of 5,000--10,000 labeled examples~\cite{liu2023zeonet}, which may not be available for a specific task or may be challenging to obtain. 
This creates a natural opportunity for self-supervised representation learning: pretrain on the full unlabeled zeolite corpus, then transfer the learned representation to downstream property-prediction tasks with limited labels.

\subsection{Self-supervised Learning for Molecular and Porous Material Representations}

Self-supervised and unsupervised representation learning have been studied extensively for molecules and materials, but most prior work has focused on graphs, atom coordinates, strings, or generative crystal representations rather than dense 3D grids. GraphMVP~\cite{liu2021pre} learns molecular representations by aligning embeddings from 2D molecular graphs and 3D coordinates. Uni-Mol~\cite{zhou2023uni} pretrains molecular representations using masked atom prediction and coordinate denoising with 3D positional encodings. LEGO~\cite{sun20243d} introduces local geometric perturbations of tetrahedral structures and trains models to reconstruct the original structure. These methods show that pretraining on geometric and chemical structure can improve molecular representation learning, but they do not directly study dense volumetric representations of porous materials.

String-based molecular pretraining provides another successful direction by training directly on molecular strings. 
ChemBERTa~\cite{chithrananda2020chemberta} and Molformer~\cite{ross2022large} adapt masked-language-modeling objectives to SMILES strings, showing that NLP-style pretraining can learn useful molecular features from large unlabeled corpora. 
Image-based approaches such as ImageMol~\cite{zeng2022accurate} instead render molecules as 2D skeletal images and train image encoders using multiple pretext tasks. 
These approaches demonstrate the flexibility of SSL across molecular representations, but their input modalities differ substantially from the 3D pore geometry and spatial fields used in zeolite-grid models.

For porous materials, several recent methods have studied representation learning and generative modeling. MOFormer~\cite{cao2023moformer} combines string-based MOFid representations with structure-based crystal graph embeddings, allowing a Transformer encoder to learn structural information without direct dense 3D inputs. Uni-MOF~\cite{wang2024comprehensive} uses atom embeddings and 3D spatial encodings in a Transformer architecture, with pretraining objectives based on masked atom prediction and coordinate denoising. Generative models such as MatterGen~\cite{zeni2023mattergen} learn distributions over atom types, coordinates, and lattice structures using diffusion-based generation. Other porous-material generative approaches have explored voxel and graph representations for reconstruction and design~\cite{naghavi2025reconstruction}. These works highlight growing interest in pretraining and generative modeling for porous materials, but they primarily operate on atomistic, graph, string, or generative structure representations.

Dense volumetric SSL has been explored in other scientific domains, especially medical image analysis, where 3D convolutional encoders are pretrained on large unlabeled image volumes~\cite{xu2025generalizable,tang2022self,wald2025openmind}. This suggests that volumetric SSL can be effective when spatial structure is central to the task. However, comparatively little work has studied SSL on dense molecular grids, such as energy grids, atomic grids, or distance grids, for porous materials, motivating this work. This gap is particularly important for zeolites, where supervised studies have already shown that dense 3D grids can be strong predictors of adsorption and synthesizability.

\section{\datasetname}\label{sec:dataset}
\datasetname is a benchmark on 3D self-supervised learning in the context of zeolites.
\datasetname consists of 402 zeolite structures from the set of experimentally synthesized structures~\cite{IZSAC} and 331,172 from the set of theoretically feasible structures~\cite{pophale2011database}.
This large set is divided into a pretraining, validation, and test set.
After pretraining, evaluation is performed on the downstream task of adsorption prediction across six different target molecules in varying label regimes for linear probing and fine-tuning. 
We release a standardized benchmarking framework and corresponding codebase for examining additional SSL methods, architectures, and augmentations with clear standardization of pretraining, training, and evaluation protocols. 
Additional details are provided below and in \S~\ref{sec:app_dataset}.

\textbf{Distance Grid Representations.}
In \datasetname, zeolites are encoded using single-channel distance grids, which have been shown to be highly effective representations for supervised learning~\cite{liu2023zeonet}.
In a signed distance grid, each voxel is assigned the distance to the nearest solvent-accessible surface formed by the framework, where negative distances lie inside the surface.
As in Figure~\ref{fig:fig1}, the distance grids of the unit cells are calculated with grid resolution 0.15 \r{A}, then are repeated and downsampled to the desired resolution $\Delta_d$ using trilinear interpolation. 
The precalculated distance grids files are provided in the ZeoNet~\cite{liu2023zeonet} codebase, while \datasetname includes dataset loading information and code.
The exact details for distance grid calculation and generation are outlined in \S~\ref{sec:app_distance_grid}.

\subsection{Training and Evaluation Protocol}\label{sec:training_eval_protocol}
\textbf{Training.} 
From the full set of 331k structures, 20k are reserved for validation and 10k are reserved for testing. 
These splits are maintained for both pretraining and evaluation on the downstream tasks, so no validation or testing data is utilized for training at any point.
During pretraining, only the pretraining set (301k structures) can be utilized for representation learning, with the goal of learning the best representations for property prediction.
The pretrained backbone is then evaluated on downstream tasks in two settings: linear probing and fine-tuning.
In linear probing, the backbone weights are frozen and a linear head is trained on the fixed embeddings of the unaugmented distance grids.
During fine-tuning, the backbone weights can be trained in tandem with the linear head on the downstream task, and training is done with augmentations.
At evaluation time, the unaugmented distance grids are inputted to the network and average performance is computed across the test set.

\textbf{Downstream Tasks.}
To evaluate representation quality, we evaluate predictions of adsorption behavior on long-chain hydrocarbon adsorption based on simulation~\cite{bai2015discovery, liu2023zeonet}.
This value is quantified by Henry's constant, $k_{\textrm{H}}$, which provides a physically meaningful metric for comparing adsorption strength across different zeolite–molecule pairs. 
Specifically, a value of ($k_{\textrm{H}} = 1$ ~mol kg$^{-1}$ MPa$^{-1}$) means that the zeolite is expected to adsorb approximately 1 mol of adsorbate per kilogram of zeolite at a pressure of 1 MPa.
Following~\cite{liu2023zeonet}, we train and evaluate using mean-squared error (MSE) of $\log(k_{\textrm{H}})$. 
The final performance is reported as MSE of the test set with bootstrapped error bars over 1000 trials.
Furthermore, to test the robustness and quality of these features, this training is done across varying training set sizes from 100 to 10,000 labeled structures for each molecule. 
These splits are constructed independently for each molecule by sampling randomly from the set of valid targets ($k_{\textrm{H}} > 0$). 
Each training set is a subset of the next largest training set.
Labels and subset splits are provided across six different hydrocarbons: three linear alkanes, ranging from C18 to C30, as well as three mono- and di-branched C18 isomers.
For further details, see \S~\ref{sec:supp_split} and \S~\ref{sec:app_adsorption}.

\section{Methods}
\subsection{Self-supervised Learning}
We perform pretraining on four self-supervised methods to learn a network $f$: BYOL~\cite{grill2020bootstrap}, DINOv1~\cite{caron2021emerging}, MoCov2+~\cite{chen2020improved}, and SimCLR~\cite{chen2020simple}.
Each method utilizes different augmentations of the same samples, which we will call $x, x'$.
Both MoCo and SimCLR are contrastive methods utilizing the InfoNCE loss~\cite{oord2018representation}:
\begin{equation}
\mathcal{L}_\text{InfoNCE} = - \log\left( \frac{\exp(f(x)^\top f(x') / \tau)}{\sum_{j}\exp(f(x)^\top f(x_j) / \tau)} \right)
\end{equation}
where $x, x'$ form a positive pair and the other samples are negatives.
SimCLR uses all other samples within the minibatch as the set of negatives while MoCo keeps a fixed-length queue of negatives which are encoded by a momentum encoder.
BYOL and DINO are distillation methods. They train a student network $f_s$ to match the output of a teacher network $f_t$, whose weights update as an exponential moving average of the student weights. 
As in contrastive learning, the student and the teacher receive different views of the same sample, but no explicit negatives are used. 
In BYOL, the student network has an additional prediction head $q$, and the objective minimizes the squared distance between the student prediction and teacher embedding:
\begin{equation}\mathcal{L}_\text{BYOL} = ||q(f_s(x)) - f_t(x')||^2_2.\end{equation}
DINO instead treats the teacher outputs as soft pseudo-labels and trains the student using cross-entropy:
\begin{equation}\mathcal{L}_\text{DINO} = H\big(\textrm{softmax} \left(c(f_t(x'))\right),\textrm{softmax}(f_s(x))\big),\end{equation}
where $c$ is a centering function, and $H(p,q)=-\sum_k p_k \log q_k$ denotes cross-entropy. To isolate the effect of the self-supervised training objective, we use the same pretraining augmentations and network architecture across methods. Specifically, all methods use the augmentations described below and a 3D ResNet-18 backbone, which was found to be effective for dense zeolite grids in~\cite{liu2023zeonet}.

\paragraph{Data Augmentations.} We primarily use geometric augmentations, since standard 2D image augmentations such as color jittering and pixel noise do not apply to dense distance grids. During SSL pretraining, we apply random 3D translations, rotations, and scale augmentations. While~\cite{liu2023zeonet} used only translations and rotations during supervised training, we find that these augmentations alone are insufficient for learning strong self-supervised representations; see \S~\ref{sec:results} for an ablation.
We implement the scale augmentation by independently varying the grid resolution along each axis:
$\Delta_d = \max\left(0.45 + \mathcal{N}(0,\sigma^2), 0.05\right)\textup{\r{A}},$
where $\sigma$ is a hyperparameter varied from $0$ (no scaling) to $0.4$. 
An example of these augmentations are shown in Figure~\ref{fig:fig1}. Scale augmentations are only used during pre-training. During linear-probe training and evaluation, we use no augmentations and fix the grid resolution to $\Delta_d = 0.45\,\textup{\r{A}}$. During fine-tuning, we use random translations and rotations also with the same fixed grid resolution, matching the setup of~\cite{liu2023zeonet}.

\input{figs_tabs/fig_baseline}
\subsection{Baselines}
We compare to baselines in both the linear probe regime and fine-tuning regime.
We focus on volumetric representations because prior work~\cite{liu2023zeonet, liu2025representation} found them to be much stronger than graph-based and coordinate-based representations for these adsorption tasks. %
We present results using other representations in Table~\ref{tab:c18_full} for completeness, but focus on handcrafted features and volumetric networks for in-depth analysis.
In terms of linear probe features, we evaluate the quality of handcrafted aggregate descriptors, 3DINO~\cite{xu2025generalizable}, and 2D networks ImageNet pretrained features. 
The handcrafted features consist of pore-limiting diameter (PLD), largest-cavity diameter (LCD), surface area, pore volume, and density of Si atoms, which are then inputted to a 2-layer MLP with hidden dimension 200 for fine-tuning. 
3DINO~\cite{xu2025generalizable} is a 3D-ViT pretrained with DINO on medical volumetric data which we evaluate for compatibility with our adsorption tasks. 
In the fine-tuning regime, we compare different pretrained backbone initializations with training ZeoNet, 3D ViT-S~\cite{dosovitskiy2020image}, and 3D Swin-S~\cite{liu2021swin} from scratch.

\textbf{Adapting ImageNet to 3D.}
We seek to compare with the quality of pretrained features from networks pretrained on ImageNet~\cite{deng2009imagenet}, but these networks are designed to take 2D images.
We utilize three methods to adapt them to 3D which we outline below and visualize in Figure~\ref{fig:fig_baseline}. 
We use volumes of shape $224 \times 224 \times D$ to be compatible with the ImageNet weights and appropriately normalize to ImageNet statistics. 
Firstly, following~\cite{liu2023zeonet} we simply select three slices from the volume to input to the network where each slice is treated as a different channel for the same image. 
We experiment with three different sampling schemes: sampling one slice at random from each spatial axis, sampling three random slices from the same axis, and taking three slices spaced at a fixed distance along the same axis.
We also propose two additional methods which utilize the full 3D volume rather than only three slices. 
In channel inflation, we duplicate the weights of the first convolutional layer to match the depth of the volume $D$, effectively treating the 3D volume as a 2D image with $D$ channels. 
To prevent initial activations from exploding, the inflated weights are downscaled by a constant hyperparameter, and in practice we found the best scale to be $\sqrt{3/D}.$
In depth tokenization, we leverage the sequential nature of vision transformers to tokenize and process the entire volume. 
We tokenize each $224 \times 224 \times 3$ chunk separately to get $16 \times 16 \times D/3$ tokens. 
These tokens are inputted sequentially to the network with only $(x, y)$ positional embeddings.
To prevent the embeddings from being excessively dependent on channel offset and molecule orientation, for all adaptations we use the average embedding of the molecule across 20 different translations and rotations.
We also tune the distance grid resolution between $\{0.15, 0.45, 0.9, 1.2\}\textup{\r{A}}.$ Exact hyperparameters are in \S~\ref{sec:supp_hyperparameters}.

\subsection{Implementation Details and Hyperparameters}\label{sec:hyperparameters}
We build on the solo-learn~\cite{solo-learn} library to conduct pretraining, linear probing, and fine-tuning experiments. 
All experiments are conducted on 3D ResNet-18 with 3D volumes of size $64 \times 64 \times 64$, other than for ImageNet networks which expect $224 \times 224$ and 3DINO which expects $112^3$.
This volume is smaller than used in ZeoNet~\cite{liu2023zeonet}, but we found comparable fine-tuning performance (Table~\ref{tab:c18_full}) while enabling higher batch sizes which are necessary for some methods such as SimCLR.
For method-level hyperparameters, we match the best hyperparameters for each method reported by solo-learn for ImageNet-100 which has a similar scale to \datasetname.
In pretraining, we use the LARS~\cite{you2017large} optimizer with batch size 1024 split between 2 NVIDIA A100 GPUs for 400 epochs, which trains for approximately 3.5 days. 
In fine-tuning and linear probe, we train using AdamW~\cite{loshchilov2017decoupled} instead of LARS on a single GPU and train for 50 and 100 epochs, respectively.
For the exact training details and configurations used in each trial, see \S~\ref{sec:supp_training}.

\vspace{-6pt}
\section{Results}\label{sec:results}
\vspace{-6pt}
\input{figs_tabs/tab_2}
\input{figs_tabs/tab_1}

\textbf{Does in-domain SSL improve low-label adsorption prediction?}
Our results show fine-tuning after pretraining leads to state-of-the-art performance, in both low and high data regimes.
In the high data regime, Table~\ref{tab:c18_full} shows the best pretraining method outperforms training from scratch on C18 adsorption prediction by 16\% on the full dataset.
This trend only becomes stronger in low data regimes.
In Figure~\ref{fig:avg_performance}, we see that as training labels decrease, the gap between pretrained and from scratch baselines becomes even wider.
We observe that SimCLR, MoCo, and BYOL outperform the state-of-the-art baselines in most subset sizes for both linear probing and fine-tuning.
Initial results in Appendix~\ref{sec:app_resnet50} suggest increased network size can lead to even further performance improvements, but the exact parameter-performance scaling laws have yet to be shown on ZeoBench.

\textbf{Which SSL objectives learn the best frozen representations?}
In Figure~\ref{fig:avg_performance} and Table~\ref{tab:adsorption_performance}, we observe that the contrastive methods SimCLR and MoCo produce the strongest features for adsorption probing, especially compared to their non-contrastive counterparts.
We speculate this is due to the more uniform distribution of embeddings learned by contrastive methods from explicit negatives~\cite{wang2020understanding}.
This can be helpful for adsorption prediction, which is a continuous regression task, in contrast to classification tasks which can benefit from clustered embeddings.
These trends appear to shift after fine-tuning; even though MoCo and SimCLR outperform BYOL by a fair margin in linear probing, BYOL is comparable in a fine-tuning setting.
Once again, this might indicate distillation methods learn useful features during pretraining but are less well-suited for regression tasks without altering the training objective. 

\input{figs_tabs/fig_baseline_results}
\textbf{How do in-domain 3D representations compare to ImageNet and medical-volume pretraining?}
Surprisingly, Figure~\ref{fig:fig_baseline_results} shows that ImageNet features are the strongest baseline for linear probing, even when having to adapt to 3D volumes.
We observe that handcrafted features are competitive with other baselines at the 10k subset but suffer in low-label regimes, likely because the limited number of descriptors can easily overfit to a small number of training points.
Despite being trained on 2D images, we find adapting 3D volumes to ImageNet pretrained networks outperforms both handcrafted features and medically pretrained features in most data regimes. 
This can be explained in part by ImageNet pretraining learning generalizable features for a broad variety of tasks, sometimes outperforming in-domain training~\cite{daroya2025wildsat}. 
However, these features are inherently limited by how much information each ``view'' captures from the 3D volume: they require many test-time augmentations to mitigate sensitivity to channel offsets and volume orientations (Figure~\ref{fig:fig_baseline_results}).

\input{figs_tabs/fig_invariance}

\textbf{Which augmentations matter for dense zeolite-grid SSL?}
In the right plot of Figure~\ref{fig:fig_baseline_results}, we see the adsorption performance for C18 is much weaker without scaling ($\sigma=0$) compared to with scaling. 
Although scaling augmentations are unnecessary for achieving strong performance when fine-tuning from scratch, they are crucial for learning generalizable features during pretraining.
We suspect with only translation and rotation, the network can learn highly local characteristics to distinguish structures rather than global ones which persist even when local shapes are warped.
This result corroborates previous work which found that stronger augmentations are more important for self-supervised learning than supervised learning~\cite{chen2020simple}.
Interestingly, in Figure~\ref{fig:fig_invariance} we see pretraining with scaling leads to lower geometric invariance but higher performance, suggesting a tradeoff between strict equivariance and more generalizable representations which transfer better to downstream tasks.

\textbf{Do transformer architectures benefit from this benchmark, or do CNNs remain stronger?}
In Figure~\ref{fig:avg_performance} and Table~\ref{tab:adsorption_performance}, we see a clear gap in performance between training ResNet from scratch compared to both ViT and Swin, corroborating prior work on this task~\cite{liu2025representation}.
One possible explanation is that the translational equivariance priors of ResNets are particularly well-suited to distance-grid prediction. 
Although prior work has suggested that trained 2D ViTs can exhibit similar or even greater translational invariance than ResNets in image classification~\cite{bouchacourt2021grounding}, it is unclear whether this behavior transfers to 3D geometric regression tasks.
In fact, Figure~\ref{fig:fig_invariance} suggests the contrary, as Swin and ViT architectures have far lower invariance after training compared to ResNets, possibly due to a relatively small dataset size being insufficient for these architectures.
This represents a clear opportunity for improvement in \datasetname, as pretraining vision transformers might offer strong gains which make them comparable or even better than CNNs on zeolite data.

\input{figs_tabs/tab_mof}

\textbf{Does ZeoBench pretraining generalize to other domains?} 
To test generalization, we conducted a transfer learning experiment on metal organic frameworks (MOFs).
While the SSL encoder operates on single-channel distance grids without reflecting chemical composition, MOFs have differing properties depending on their transition-metal nodes (Cu, Zn, Zr, etc.) and organic linkers (C/N/O). 
We conducted an initial exploration on an entirely out-of-domain transfer learning by fine-tuning our ZeoBench SimCLR pretrained backbone on the hMOF dataset~\cite{wilmer2012large}. 
We constructed distance grids for hMOFs by assigning each atom its van der Waals radius. 
Despite this gap, in Table~\ref{tab:mof} we find that ZeoBench pretraining consistently outperforms existing methods and training from scratch, especially for low-data regimes, indicating the features learned in pretraining transfer effectively to an unseen domain.

\vspace{-6pt}
\section{Limitations and Future Work} \label{sec:limitations}
\vspace{-6pt}

Despite the strong performance of the self-supervised methods investigated, we see clear room to improve and expand both our analysis and the broader \datasetname benchmark. 
Our current study focuses primarily on a fixed 3D ResNet-18 backbone to isolate the effect of the pretraining objective. 
Future work should evaluate a broader range of architectures and self-supervised methods, including transformer-specific reconstruction objectives~\cite{he2022masked} and group-equivariant networks~\cite{cohen2016group, xu2023eequivariant, tai2019equivariant} that incorporate rotation and translation symmetries.
Group-equivariant models are particularly appealing for 3D zeolite grids, but are challenging to scale because activation memory and compute typically grow approximately linearly with the group size $K$ and representing rotations well in 3D may require many group elements.
More broadly, we have not yet characterized scaling relationships with larger backbones, longer pretraining, larger batch sizes, or more compute. 
Such studies would clarify whether trends from 2D vision transfer to zeolite representation learning, and whether the current advantage of CNNs over vision transformers persists at larger scale.

We also hope to expand the set of frameworks, target molecules, and downstream tasks included in \datasetname as simulated and experimental data become available. 
The current benchmark focuses on adsorption prediction for long-chain hydrocarbons in all-silica zeolites, but the same framework naturally extends to other separation and catalysis-relevant tasks. 
More broadly, future versions of \datasetname could support models that jointly encode porous structures and target molecules to predict their interactions across zeolites, metal-organic frameworks, and other porous materials. 
Further visualization and analysis of learned feature spaces could also help reveal which geometric or topological properties are captured by self-supervised representations.

\vspace{-6pt}
\section{Conclusion}\label{sec:conclusion}
\vspace{-6pt}
Despite the rigorous studies conducted on zeolite structures in the context of supervised learning, a large majority of these structures remain unutilized due to expensive or lack of applicable labels.
To address this gap, we present \datasetname, a benchmark for self-supervised representation learning in the context of volumetric computer vision evaluated on six downstream adsorption prediction tasks for different training and label regimes. 
We find that self-supervised pretraining not only leads to state-of-the-art performance in fine-tuning and excels in low-label regimes, but extends to out-of-domain improvements on metal organic frameworks. 
Our analysis also reveals that contrastive pretraining outperforms distillation-based methods for adsorption prediction and suggests transformer architectures require additional pretraining to match CNN performance.
We see clear room for improvement on \datasetname, particularly in terms of expanding pretraining methodologies, architectures, and downstream tasks.
We hope that \datasetname can aid in the future discovery of synthesizable zeolites for furthering the efficiency of chemical industrial applications.

\section*{Acknowledgements}  
The project was supported by the National Science Foundation under Award Number ACED-2435659. The research used resources through the NAIRR pilot compute grant 250024 and of the Advanced Cyberinfrastructure Coordination Ecosystem: Services \& Support (ACCESS) through Allocation Nos. CHM250039 and CHM250051.

{
\small
\bibliographystyle{plainnat}
\bibliography{main}
}

\input{appendix}

\newpage
\input{checklist.tex}

\end{document}

%% file: figs_tabs/fig_1.tex
\begin{figure}[ht]
    \centering
    \includegraphics[width=0.93\linewidth]{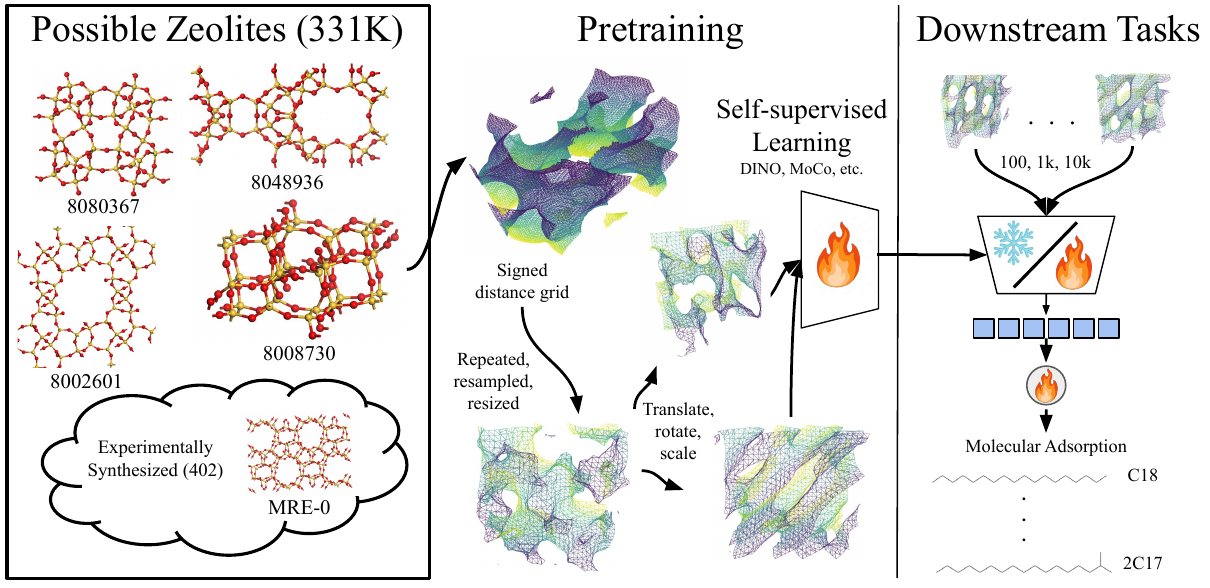}
    \vspace{-3pt}
    \caption{
        \textbf{\datasetname pretraining and evaluation setup.}
        The full set of theoretical and experimentally synthesized zeolites are used for pretraining and evaluation.
        Translation, rotation, and scaling are used as augmentations for self-supervised learning on a signed distance grid.
        The level set of zero is plotted with color showing depth.
        After pretraining, the backbone is linearly probed (snowflake) or fine-tuned (flame) on six different molecular adsorption tasks with varying subset sizes (100 to 10k).
    }
    \vspace{-20pt}
    \label{fig:fig1}
\end{figure}

%% file: figs_tabs/fig_baseline.tex
\begin{figure}[h]
    \centering
    \includegraphics[width=0.95\linewidth]{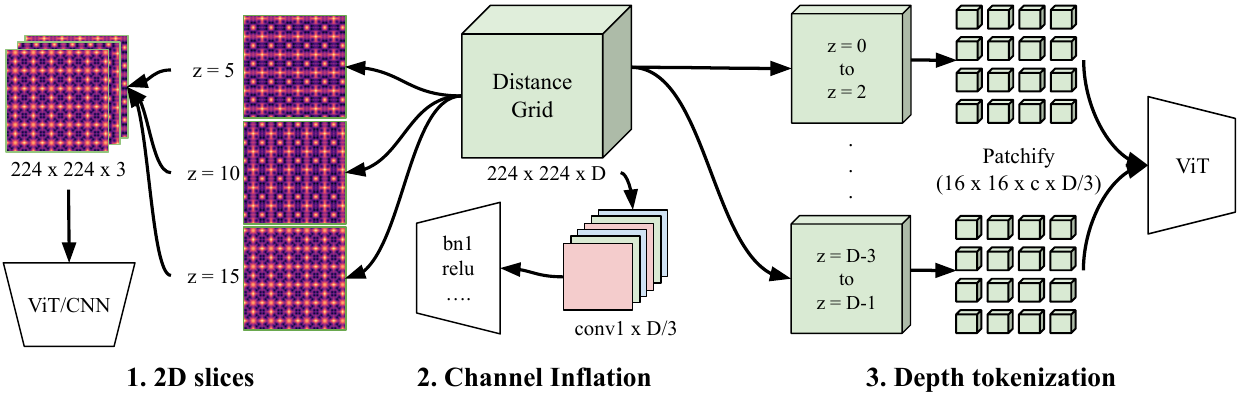}
    \vspace{-6pt}
    \caption{
        \textbf{Adapting 2D networks for 3D inputs.}
        We explore three methods to get ImageNet features from volumetric inputs: 2D slices, channel inflation, and depth tokenization.
        2D slices (left) involve stacking three different slices to form an image.
        Channel inflation (center) repeats the weights of the first convolutional layer.
        Depth tokenization (right) treats the volume as a stack of $D/3$ images.
    }
    \vspace{-6pt}
    \label{fig:fig_baseline}
\end{figure}

%% file: figs_tabs/tab_2.tex
\begin{figure}[h]
    \vspace{-6pt}
    \centering
    \begin{minipage}{0.65\textwidth}
        \centering
        \includegraphics[width=\linewidth]{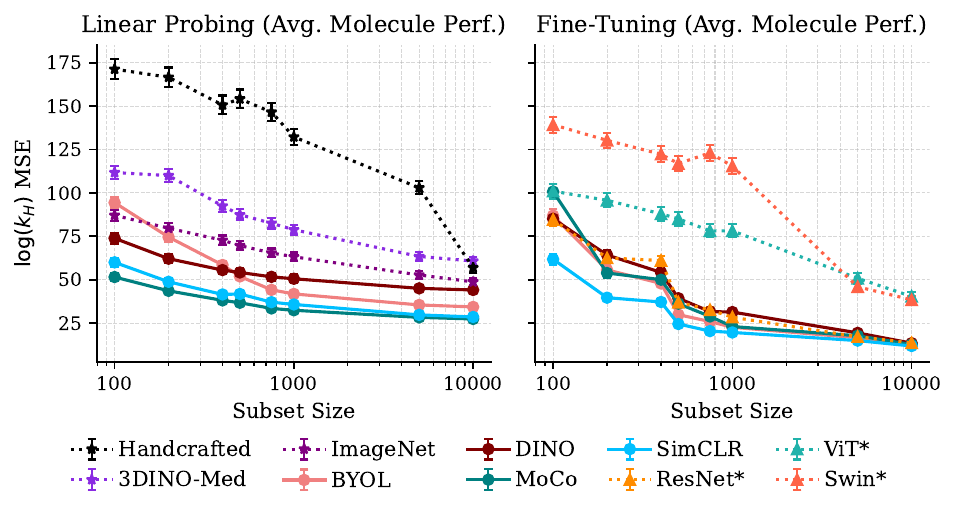}
        \caption{
        \textbf{Method performance on \datasetname averaged across all six downstream tasks for varying subset sizes (lower is better).}
        For linear probe (left), SimCLR and MoCo outperform other methods in all contexts.
        For fine-tuning (right), SSL methods (solid) outperform training from scratch (dotted), especially when labeled data is limited.
        ImageNet results shown are the best-performing adaptation method, depth tokenization.
        }
        \label{fig:avg_performance}
    \end{minipage}\hfill %
    \begin{minipage}{0.33\textwidth}
        \centering
        
        \resizebox{\linewidth}{!}{
        \tiny
        \begin{tabular}{l | c}
            \toprule
            Method & C18 Full ($\downarrow$) \\
            \midrule
            M3GNet* & 18.5 $\pm$ ---\\
            EdgeConv* & 65.1 $\pm$ ---\\
            ZeoNet & 3.7 $\pm$ 0.2 \\
            \midrule
            BYOL & 3.6 $\pm$ 0.2 \\
            DINO & \textit{3.2 $\pm$ 0.2} \\
            MoCo & \textbf{3.1 $\pm$ 0.2} \\
            SimCLR & 3.3 $\pm$ 0.2 \\
            \bottomrule
        \end{tabular}
        }
        \vspace{-3pt} %
        \captionof{table}{
        \textbf{Adsorption performance on C18 full training set (70k).}
        * = reported in prior work~\cite{liu2025representation}.
        M3GNet~\cite{chen2022universal} uses a graph network while EdgeConv~\cite{wang2019dynamic} uses a coordinate representation.
        SSL pretraining (bottom) leads to stronger performance than baselines under an identical training setup.
        }
        \label{tab:c18_full}
    \end{minipage}
    \vspace{-6pt}
\end{figure}

%% file: figs_tabs/tab_1.tex
\begin{table}[h]
    \centering
    \resizebox{\textwidth}{!}{
    \begin{tabular}{ll | cccccc | c}
        \toprule
        Pretrain & Arch & C18 & C24 & C30 & 2C17 & 4C17 & 22C16 & Average \\
        \midrule
        \multicolumn{9}{c}{\textit{Linear Probing}} \\
        \midrule
        ImageNet & ViT-2D & 41.7 $\pm$ 0.9 & 57.0 $\pm$ 1.3 & 82.1 $\pm$ 1.9 & 38.4 $\pm$ 0.8 & 38.2 $\pm$ 0.8 & 35.1 $\pm$ 0.8 & 48.7 $\pm$ 1.1 \\
        Hand & MLP & 44.7 $\pm$ 1.8 & 60.6 $\pm$ 1.4 & 98.4 $\pm$ 2.1 & 43.5 $\pm$ 1.7 & 48.1 $\pm$ 1.4 & 46.2 $\pm$ 1.9 & 56.9 $\pm$ 1.7 \\
        3DINO~\cite{xu2025generalizable} & ViT & 56.5 $\pm$ 1.1 & 71.9 $\pm$ 1.5 & 93.2 $\pm$ 2.0 & 48.9 $\pm$ 1.0 & 48.5 $\pm$ 0.9 & 45.5 $\pm$ 0.9 & 60.8 $\pm$ 1.2 \\
        
        BYOL & ResNet & 27.4 $\pm$ 0.7 & 38.0 $\pm$ 0.9 & 65.6 $\pm$ 1.5 & 24.8 $\pm$ 0.6 & 25.2 $\pm$ 0.6 & 24.8 $\pm$ 0.6 & 34.3 $\pm$ 0.8 \\
        DINO & ResNet & 39.4 $\pm$ 0.9 & 50.8 $\pm$ 1.3 & 77.5 $\pm$ 1.9 & 33.4 $\pm$ 0.8 & 32.7 $\pm$ 0.8 & 30.6 $\pm$ 0.8 & 44.1 $\pm$ 1.1 \\
        MoCo & ResNet & \textbf{21.5 $\pm$ 0.6} & \textbf{29.7 $\pm$ 0.9} & \textbf{58.7 $\pm$ 1.6} & \textbf{18.4 $\pm$ 0.5} & \textbf{18.6 $\pm$ 0.6} & \textbf{17.3 $\pm$ 0.5} & \textbf{27.4 $\pm$ 0.8} \\
        SimCLR & ResNet & \textit{22.2 $\pm$ 0.6} & \textit{30.7 $\pm$ 0.9} & \textit{59.5 $\pm$ 1.5} & \textit{19.8 $\pm$ 0.5} & \textit{19.8 $\pm$ 0.5} & \textit{19.1 $\pm$ 0.6} & \textit{28.5 $\pm$ 0.8} \\
        \midrule
        \multicolumn{9}{c}{\textit{Fine-Tuning}} \\
        \midrule
        Scratch & ResNet & 6.2 $\pm$ 0.3 & 12.8 $\pm$ 0.4 & 41.8 $\pm$ 1.4 & 6.1 $\pm$ 0.2 & 7.1 $\pm$ 0.3 & 7.2 $\pm$ 0.3 & 13.5 $\pm$ 0.5 \\
        Scratch & Swin & 31.1 $\pm$ 1.1 & 43.1 $\pm$ 1.4 & 76.4 $\pm$ 2.2 & 27.7 $\pm$ 0.9 & 26.6 $\pm$ 0.9 & 25.5 $\pm$ 0.9 & 38.4 $\pm$ 1.2 \\
        Scratch & ViT & 31.3 $\pm$ 1.1 & 47.4 $\pm$ 1.7 & 80.8 $\pm$ 2.3 & 27.8 $\pm$ 0.9 & 27.0 $\pm$ 1.0 & 27.9 $\pm$ 1.1 & 40.4 $\pm$ 1.3 \\
        
        BYOL & ResNet & 5.5 $\pm$ 0.2 & \textit{11.6 $\pm$ 0.4} & \textit{38.7 $\pm$ 1.4} & \textit{5.7 $\pm$ 0.2} & \textit{6.7 $\pm$ 0.3} & 6.5 $\pm$ 0.2 & \textit{12.4 $\pm$ 0.5} \\
        DINO & ResNet & 6.3 $\pm$ 0.2 & 12.6 $\pm$ 0.4 & 41.2 $\pm$ 1.3 & 6.0 $\pm$ 0.2 & 7.2 $\pm$ 0.3 & 7.1 $\pm$ 0.2 & 13.4 $\pm$ 0.4 \\
        MoCo & ResNet & \textit{5.4 $\pm$ 0.2} & 11.7 $\pm$ 0.4 & 39.3 $\pm$ 1.3 & 6.1 $\pm$ 0.2 & 6.8 $\pm$ 0.3 & \textit{6.3 $\pm$ 0.2} & 12.6 $\pm$ 0.4 \\
        SimCLR & ResNet & \textbf{4.9 $\pm$ 0.2} & \textbf{10.5 $\pm$ 0.4} & \textbf{37.7 $\pm$ 1.4} & \textbf{5.2 $\pm$ 0.2} & \textbf{6.3 $\pm$ 0.3} & \textbf{6.0 $\pm$ 0.2} & \textbf{11.8 $\pm$ 0.4} \\
        \bottomrule
    \end{tabular}
    }
    \vspace{3pt}
    \caption{
    \textbf{Adsorption Performance ($\log(k_H)$ MSE, lower is better) across molecules (10k subset).}
    Best results are given in \textbf{bold} and second-best are given in \textit{italics}.
    Overall, SimCLR and MoCo tend to perform best in linear probe but BYOL becomes comparable after fine-tuning. 
    Self-supervised learning methods generally outperform baselines.
    Full results are given in \S~\ref{sec:app_full_results}.
    }
    \vspace{-6pt}
    \label{tab:adsorption_performance}
\end{table}

%% file: figs_tabs/fig_baseline_results.tex
\begin{figure}[h]
    \centering
    \includegraphics[width=0.99\linewidth]{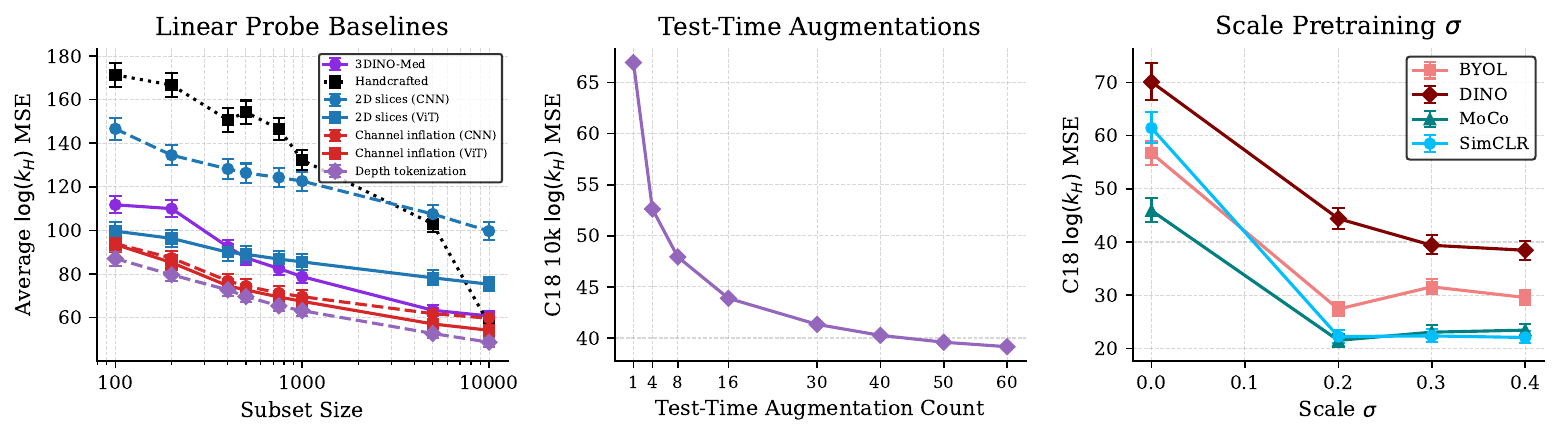}
    \vspace{-6pt}
    \caption{
        \textbf{Ablation studies.}
        Left: 
        Comparison of linear probe baseline average performance on \datasetname tasks. 
        Surprisingly, ImageNet pretraining with depth tokenization is the best-performing baseline.
        Center: 
        Effects of test-time augmentation on best ImageNet adaptation method, depth tokenization.
        Adapting 2D networks to 3D inputs requires significant test-time augmentation to achieve invariance to channel offsets and molecule rotation.
        Right: 
        Linear probe performance on C18 10k after pretraining with varying scale noise $\sigma$.
        While the presence of scaling in pretraining is crucial for strong performance on adsorption, the degree of augmentation strength seems to have little impact on performance.
    }
    \label{fig:fig_baseline_results}
\end{figure}

%% file: figs_tabs/fig_invariance.tex
\begin{figure}[h]
    \centering
    \includegraphics[width=0.99\linewidth]{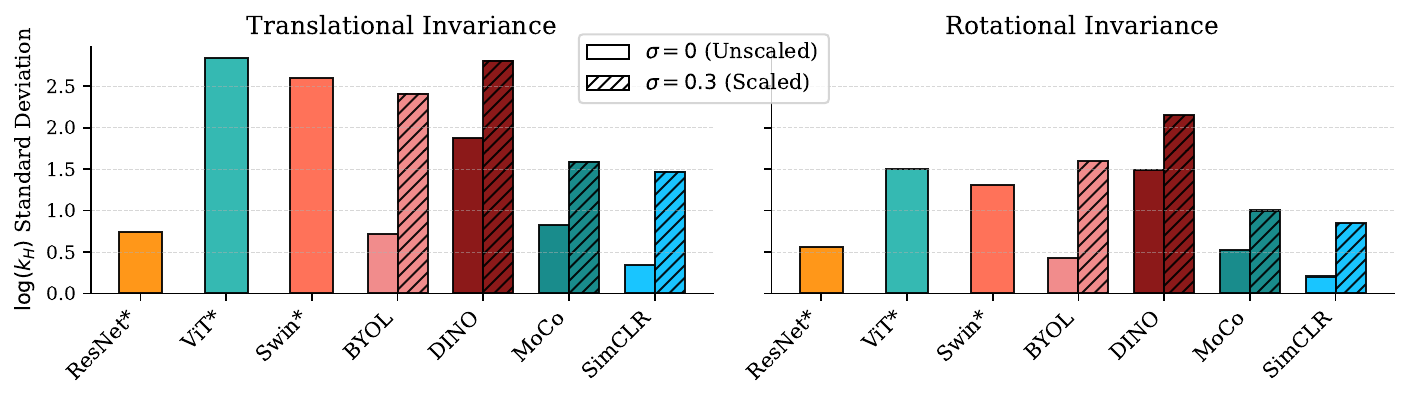}
    \vspace{-6pt}
    \caption{
        \textbf{Rotational and translational invariance of trained networks for the C18 10k subset (lower is better).}
        * means the model is trained from scratch, while other models are pretrained and linear probed. 
        Invariance is calculated as the standard deviation in predicted $\log k_{\textrm{H}}$ across 100 augmented views of a single structure, averaged across the test set.
        Although networks pretrained without scale are more invariant, they perform worse on the adsorption prediction task (Figure~\ref{fig:fig_baseline_results}). 
        Transformer architectures show far less translational invariance than ResNets even after training, likely because of equivariant properties inherent to ResNets.
    }
    \vspace{-12pt}
    \label{fig:fig_invariance}
\end{figure}

%% file: figs_tabs/tab_mof.tex
\begin{table}[h]
    \centering
    \resizebox{\textwidth}{!}{
    \begin{tabular}{lll | cccccc | c}
        \toprule
        Train Size & Pretrain & Arch & $\text{CO}_2$ 0.05 bar & $\text{CO}_2$ 0.5 bar & $\text{CO}_2$ 2.5 bar & $\text{CH}_4$ 0.05 bar & $\text{CH}_4$ 0.5 bar & $\text{CH}_4$ 2.5 bar & Average \\
        \midrule
        1,000 & Scratch & ResNet-18 & 0.187 & 0.601 & 1.108 & 0.040 & 0.175 & 0.460 & 0.429 \\
        1,000 & SimCLR & ResNet-18 & 0.180 & 0.537 & 0.922 & 0.037 & 0.175 & 0.408 & 0.377 \\
        \midrule
        5,000 & Scratch & ResNet-18 & 0.159 & 0.520 & 0.932 & 0.035 & 0.170 & 0.435 & 0.375 \\
        5,000 & SimCLR & ResNet-18 & 0.155 & 0.348 & 0.588 & 0.034 & 0.116 & 0.211 & 0.242 \\
        \midrule
        96,356 & Barlow Twins & MOFormer\cite{cao2023moformer} & 0.110 & 0.330 & 0.645 & 0.022 & 0.099 & 0.239 & 0.241 \\
        96,356 & Scratch & ResNet-18 & 0.072 & 0.185 & 0.315 & 0.013 & 0.051 & 0.108 & 0.124 \\
        96,356 & SimCLR & ResNet-18 & 0.067 & 0.182 & 0.309 & 0.013 & 0.048 & 0.106 & 0.121 \\
        \bottomrule
    \end{tabular}
    }
    \vspace{3pt}
    \caption{
    \textbf{Mean absolute error (mol $\text{kg}^{-1}$) of gas adsorption prediction for varying molecules and training sizes on the hMOF dataset~\cite{wilmer2012large}.}
    This table matches the results in Table 2 in the work presenting MOFormer~\cite{cao2023moformer}, a MOF-specific architecture which pretrains on SMILES text strings. 
    We observe that a SimCLR pretrained model on ZeoBench outperforms existing baselines and training from scratch even on a new task, indicating the features learned during pretraining generalize to a new domain. 
    This is most apparent at smaller training sets but persists even for the full training set.
    }
    \vspace{-6pt}
    \label{tab:mof}
\end{table}

%% file: appendix.tex
\newpage
\appendix

\setcounter{table}{0}   
\renewcommand{\thetable}{A\arabic{table}}
\setcounter{figure}{0}
\renewcommand{\thefigure}{A\arabic{figure}}

\section{Appendix}
\subsection{\datasetname}
\label{sec:app_dataset}
\datasetname consists of a codebase and standardized framework for evaluating and analyzing self-supervised learning methods for zeolite adsorption prediction.
We build upon the solo-learn~\cite{solo-learn} codebase for modularity of architectures, pretraining methodology, and evaluation. 
We release \datasetname under the CC BY-NC-SA 4.0 license.

\subsubsection{Motivation and Background}
\datasetname is a benchmarking dataset on zeolites, a class of nanoporous molecules consisting solely of corner-sharing SiO\textsubscript{4} tetrahedra. 
These tetrahedra can combine into different repeating structures called unit cells to create stable crystal structures.
Despite always sharing the same chemical composition, the shape selectivity of these structures vary greatly depending solely on the spatial arrangement of these frameworks~\cite{smit2008molecular}.
For instance, for the same target molecule two zeolites can vary in adsorption by orders of magnitude depending on specific pore structures. 
Hence, the problem of framework-solvent adsorption is an excellent application of computer vision, since it is fundamentally a problem of 3D geometry as opposed to specific chemical interactions.
In fact, existing works have found that in supervised settings, 3D CNNs can predict this property with very similar accuracy and substantially lower runtime compared to simulations~\cite{liu2023zeonet}.

Thus far, supervised learning has only been performed on a small subset of the zeolites.
Although experimental results have shown adsorption labels can be accurately predicted through Monte Carlo simulations~\cite{bai2013trappe, bai2015discovery, bai2014understanding}, the computational demands of these simulations are significant, taking about one hour per structure-molecule pair, and a large portion of structures can be inapplicable for a given task~\cite{liu2023zeonet}. 
Furthermore, these simulations must be run entirely from scratch when considering a new target molecule. 
Ideally, we want to utilize the full training data to produce generalizable representations which are applicable to different downstream tasks with limited labeled data.
To evaluate this, we present \datasetname, a benchmark focused on 3D self-supervised learning in the context of zeolites.
By utilizing the full range of 331,574 structures, we evaluate the representations learned through various pretraining methods on six different adsorption prediction tasks with varying numbers of labeled structures. 
We release a benchmarking framework and corresponding codebase for experimenting with additional SSL methods, architectures, and analyses with clear standardization of pretraining, training, and evaluation protocols. 

\subsubsection{Dataset Split Information}\label{sec:supp_split}
\input{figs_tabs/supp_tab_split}
\datasetname provides a total of 8 different subset sizes across 6 different molecular downstream tasks.
The validation set and test set match those of ZeoNet~\cite{liu2023zeonet} and the remaining structures make up the pretraining set. 
Subsets of varying sizes are utilized for downstream training evaluation: 100, 200, 400, 500, 750, 1000, 5000, 10000. 
Each smaller set is a subset of the next largest set.
For example, the structures in the size 100 subset are all in the size 200 subset, and so on. 
These subsets are selected independently at random for each target molecule from the set of valid structures for adsorption. 
This creates 48 different training splits for downstream evaluation. 
We also perform evaluations on the full training set from ZeoNet in Table~\ref{tab:c18_full}.
The exact split sizes are given in Table~\ref{tab:supp_splits}.

\subsubsection{Distance Grid Data}\label{sec:app_distance_grid}
For each structure, we calculate the distance grids of the unit cell using Zeo++ version 0.3~\cite{willems2012algorithms} with probe radius 1.2 \r{A} and grid resolution 0.15 \r{A}, where Si and O atoms have radii of 2.1\r{A} and 1.52\r{A}, respectively.
To generate the volumes used in training, the unit cell is repeated, then downsampled at the desired resolution $\Delta_d$ using trilinear interpolation. 
This process is shown in Figure~\ref{fig:fig1}.
The precalculated distance grids files are provided in the ZeoNet~\cite{liu2023zeonet} codebase, while \datasetname includes dataset loading information and code.

\subsubsection{Downstream Adsorption Property Prediction}\label{sec:app_adsorption}
The primary downstream task examined in \datasetname is structure-target adsorption behavior for six long-chain hydrocarbons.
More concretely, for a given hydrocarbon structure, the goal is to predict Henry's constant, $k_{\textrm{H}}$, which describes the low-pressure adsorption affinity for a given zeolite-molecule pair. 
This value quantifies the rate of adsorption for this pair: a value of ($k_{\textrm{H}} = 1$ ~mol kg$^{-1}$ MPa$^{-1}$) means that the zeolite is expected to adsorb approximately 1 mol of adsorbate per kilogram of zeolite at a pressure of 1 MPa.
Labels are provided for six hydrocarbon molecules: C18, C24, C30, 2C17, 4C17, 22C16. 

\subsection{Training Information}\label{sec:supp_training}
\subsubsection{Hyperparameter Sweep}
We use a cosine decay learning rate schedule after a 10 epoch warmup. 
We tune the learning rate between $\{10, 1, 0.1, 0.01\}$ but in practice we found $1$ to be the best learning rate for all pretraining.
Weight decay is set as $10^{-6}$. 
In fine-tuning and linear probe, we train using AdamW~\cite{loshchilov2017decoupled} instead of LARS on a single GPU.
Fine-tuning is run for 50 epochs with 5 warmup epochs while linear probing is run for 100 epochs with 10 warmup epochs with batch size 128 in both cases.
For each linear-probe trial, we tune the learning rate and weight decay on the validation set using grid search over
$\{10^{-1}, 10^{-2}, 10^{-3}\}$ and $\{10^{-2}, 10^{-3}, 10^{-4}\}$, respectively.
For fine-tuning, hyperparameters are tuned on the 10k subset C18 adsorption validation performance and fixed for the other subsets and tasks.
We utilize a smaller batch size of 16 for ViT and Swin training due to higher memory requirements. 
The best Swin hyperparameters were patch size 4 and window size 9, while the best ViT hyperparameters found were using sincos positional embeddings and patch size 8. 
In all experiments, the precision used is 16-mixed. 

In pretraining, we use the LARS~\cite{you2017large} optimizer with batch size 1024 split between 2 NVIDIA A100 GPUs for 400 epochs, which trains for approximately 3.5 days. 
In fine-tuning and linear probe, we train using AdamW~\cite{loshchilov2017decoupled} instead of LARS on a single GPU and train for 50 and 100 epochs, respectively.
This takes about 10 minutes for the linear probe and 12 hours for the fine-tuning.

\subsubsection{Pretraining Hyperparameters} \label{sec:supp_hyperparameters}
BYOL~\cite{grill2020bootstrap} was run with a projection dimension of 4096, prediction dimension of 8192, and output dimension of 256. 
DINO~\cite{caron2021emerging} was run with a hidden dimension of 2048 and output dimension of 256. The number of prototypes used is 4096. Finally, the training was also done with 50 warmup epochs for the teacher temperature and gradients were clipped to be 3.0 or less based on stability issues in training.
MoCoV2+~\cite{chen2020improved} was run with a hidden dimension of 2048 and output dimension 256, a queue size of 65536 and temperature 0.2. 
SimCLR was run with a hidden dimension of 4096, output dimension 256, and temperature 0.2.

For ImageNet baselines, the three different 2D slice sampling schemes were done but the best performing one was equidistant sampling with distance 3 and resolution 0.45. 
The channel inflation method was swept across 30, 60, and 90 channels but 30 channels performed the best with a resolution of 0.15. 
The best depth tokenization settings found were with depth 30 and resolution 0.45. 

\subsection{Additional Results}
\subsubsection{ResNet-50 Results}\label{sec:app_resnet50}
\input{figs_tabs/tab_resnet50}
As an extension, we tested the parameter scaling properties of our best-performing setup, SimCLR, by running identical pretraining followed by linear probing and fine-tuning experiments on ResNet-50. 
The results are shown in Table~\ref{tab:adsorption_performance_resnet50}.
We find that the larger network (ResNet-50, 57M parameters) outperforms the smaller network (ResNet-18, 33M parameters) across all settings. 
We suggest that these trends could be explored more thoroughly in future work as to how these scale further with larger networks and whether these trends hold across different methods.

\subsubsection{IZSAC Subset Results}\label{sec:app_izsac_results}
\input{figs_tabs/tab_izsac}
We also present results for our best-performing model on the IZSAC subset alone. 
Since PCOD and IZSAC molecules were split uniformly between training, validation, and testing, the test set only contains 20 structures. 
In Tables~\ref{tab:izsac_10k} and \ref{tab:izsac_full} we compare the average performance on the full test set, the IZSAC subset of the test set, and the IZSAC subset of the validation and test sets combined.
We see that the overall trends follow the full test set although the fewer data points mean the results are noisier. 

\subsubsection{Feature Visualizations}\label{sec:app_umap}
\input{figs_tabs/fig_umap}
For our best performing model (SimCLR pretrained on $\sigma=0.2$), we tried visualizing the UMAP~\cite{mcinnes2018umap} embedding projections alongside the handcrafted features which signify important molecular properties. 
These are shown in Figure~\ref{fig:app_umap}. 
We also visualized synthesizability, or the difference in IZSAC zeolites compared to PCOD zeolites, but we found minimal correlation given the low amount of IZSAC zeolites in the visualization. 

\subsubsection{Full Results Tables}\label{sec:app_full_results}
\input{figs_tabs/tab_full}
In addition to the fine-tuning on the full C18 training set in Table~\ref{tab:c18_full}, we present results in Table~\ref{tab:app_full_train} on fine-tuning our best self-supervised model along on the full training sets available for the other molecules. 
We observe similar performance improvements from pretraining across all six downstream tasks. 
The full downstream task performance results are given in Tables~\ref{tab:adsorption_subsets_c18}, \ref{tab:adsorption_subsets_c24}, \ref{tab:adsorption_subsets_c30}, \ref{tab:adsorption_subsets_2c17}, \ref{tab:adsorption_subsets_4c17}, \ref{tab:adsorption_subsets_22c16} with the average performance given in Table~\ref{tab:adsorption_subsets_average}.
\input{figs_tabs/tab_all_results}

%% file: figs_tabs/supp_tab_split.tex
\begin{table}[h]
    \centering
    \begin{tabular}{l | c}
        \toprule
        Split & Subset Size \\
        \midrule
        Pretrain & 301470\\
        Validation & 20104\\
        Test & 10000\\
        Downstream Train & 100, 200, 400, 500, 750, 1000, 10000 \\
        \bottomrule
    \end{tabular}
    \vspace{3pt}
    \caption{
    \textbf{Pretraining, training, validation, and test splits for \datasetname.}
    Validation and test sets match ZeoNet~\cite{liu2023zeonet}.
    Downstream training sets are subsets of the pretraining set and each set is a subset of the next largest set.
    }
    \label{tab:supp_splits}
\end{table}

%% file: figs_tabs/tab_resnet50.tex
\begin{table}[!h]
    \centering
    \resizebox{\textwidth}{!}{
    \begin{tabular}{ll | cccccc | c}
        \toprule
        Pretrain & Arch & C18 & C24 & C30 & 2C17 & 4C17 & 22C16 & Average \\
        \midrule
        \multicolumn{9}{c}{\textit{Linear Probing}} \\
        \midrule
        SimCLR & ResNet-18 & 22.3 & 31.4 & 61.4 & 19.9 & 19.9 & 19.5 & 29.1 \\
        SimCLR & ResNet-50 & 19.9 & 26.5 & 53.3 & 17.1 & 17.2 & 16.8 & 25.1\\
        \midrule
        \multicolumn{9}{c}{\textit{Fine-Tuning}} \\
        \midrule
        SimCLR & ResNet-18 & 5.2 & 10.3 & 39.8 & 5.0 & 6.7 & 6.2 & 12.2 \\
        SimCLR & ResNet-50 & 3.1 & 9.2 & 28.1 & 4.8 & 6.0 & 6.0 & 9.5 \\
        \bottomrule
    \end{tabular}
    }
    \vspace{3pt}
    \caption{
    \textbf{Effects of parameter scaling on SSL downstream performance for SimCLR (10k subset).}
    Table format and results match Table~\ref{tab:adsorption_performance}.
    Running identical pretraining followed by linear probing and fine-tuning on ResNet-50 yields improved results compared to ResNet-18. 
    }
    \label{tab:adsorption_performance_resnet50}
\end{table}

%% file: figs_tabs/tab_izsac.tex
\begin{figure}[h]
    \vspace{-6pt}
    \centering
    \begin{minipage}{0.6\textwidth}
        \centering
        \resizebox{\textwidth}{!}{
        \begin{tabular}{l | cccccc | c}
            \toprule
            ZeoBench Subset & C18 & C24 & C30 & 2C17 & 4C17 & 22C16 & Average \\
            \midrule
            \multicolumn{8}{c}{\textit{Linear Probing}} \\
            \midrule
            Full Test Set & 22.2 & 30.7 & 59.5 & 19.8 & 19.8 & 19.1 & 28.5 \\
            IZSAC Test (20) & 30.5 & 21.4 & 79.9 & 22.1 & 31.4 & 23.2 & 34.8 \\
            IZSAC Val+Test (76) & 27.7 & 32.2 & 66.8 & 17.8 & 22.5 & 20.8 & 31.3 \\
            \midrule
            \multicolumn{8}{c}{\textit{Fine-Tuning}} \\
            \midrule
            Full Test Set & 4.9 & 10.6 & 37.7 & 5.2 & 6.3 & 6.0 & 11.8 \\
            IZSAC Test (20) & 6.0 & 10.7 & 38.4 & 7.6 & 3.7 & 7.4 & 12.3 \\
            IZSAC Val+Test (76) & 7.3 & 10.3 & 20.9 & 4.5 & 5.2 & 5.9 & 9.0 \\
            \bottomrule
        \end{tabular}
        }
        \vspace{3pt}
        \captionof{table}{
        \textbf{Adsorption Performance ($\log(k_H)$ MSE, lower is better) of SimCLR pretrained model with 10k subset pretraining on experimentally-verified (IZSAC) structures.}
        }
        \label{tab:izsac_10k}
    \end{minipage}\hfill %
    \begin{minipage}{0.38\textwidth}
        \centering
        
        \resizebox{\linewidth}{!}{
        \tiny
        \begin{tabular}{l | c}
            \toprule
            Zeolite Subset & C18 MSE ($\downarrow$) \\
            \midrule
            Full Test & 3.3 \\
            IZSAC Test (20) & 4.3 \\
            IZSAC Val + Test (76) & 5.7 \\
            \bottomrule
        \end{tabular}
        }
        \vspace{-3pt} %
        \captionof{table}{
        \textbf{Adsorption performance of SimCLR pretrained model on C18 full training set (70k) for full test set and IZSAC molecules only.}
        }
        \label{tab:izsac_full}
    \end{minipage}
    \vspace{-6pt}
\end{figure}

%% file: figs_tabs/fig_umap.tex
\begin{figure}[h]
    \centering
    \includegraphics[width=0.9\linewidth]{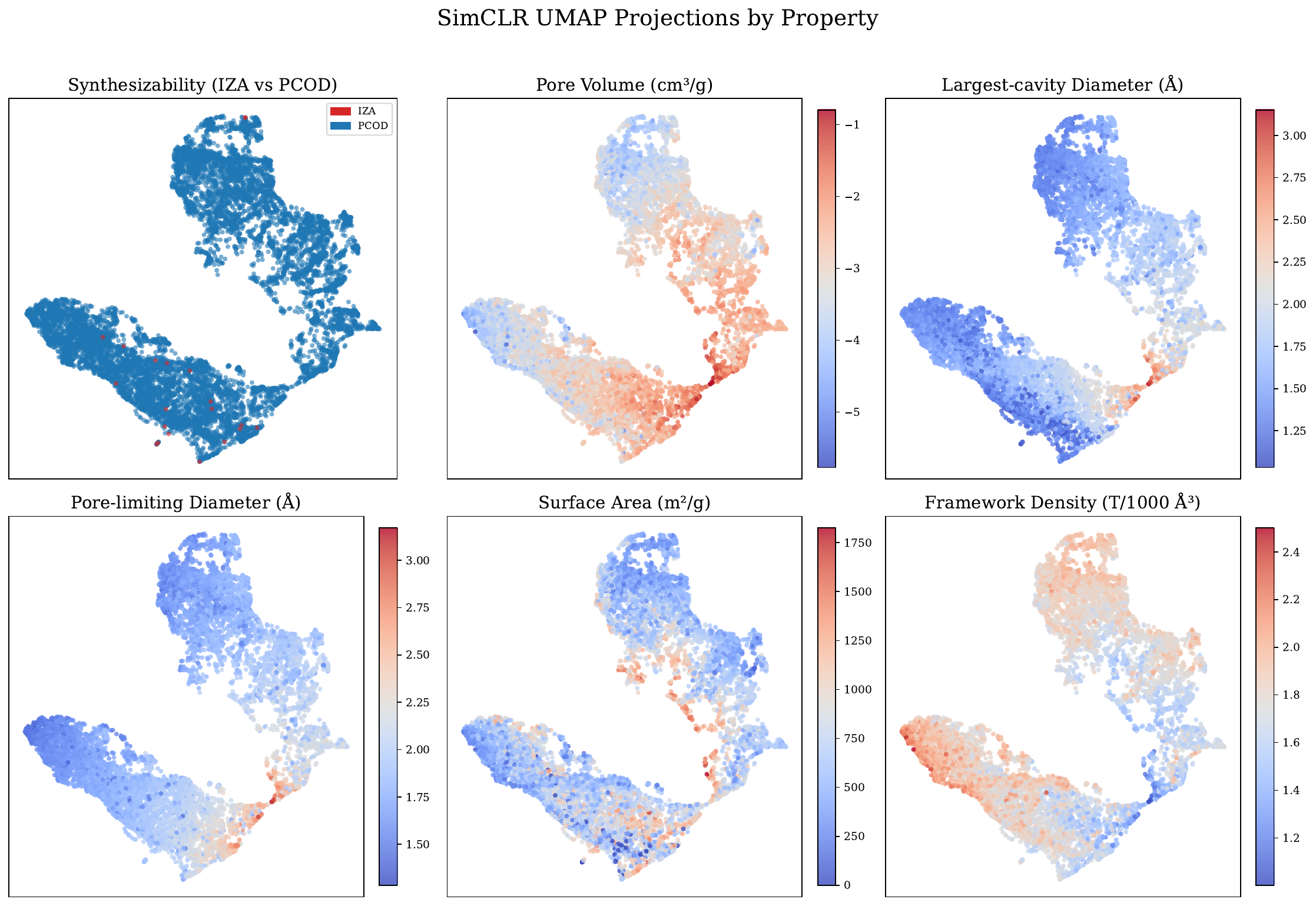}
    \vspace{-6pt}
    \caption{
        \textbf{UMAP visualizations for our best performing pretrained features alongside handcrafted features and synthesizability.}
        We see the pretrained features are fairly strong predictors of most handcrafted features like pore volume and framework density. 
        Synthesizability prediction (IZSAC vs PCOD), on the other hand, does not seem to be predicted by the UMAP projections. 
    }
    \label{fig:app_umap}
\end{figure}

%% file: figs_tabs/tab_full.tex
\begin{table}[!h]
    \centering
    \resizebox{\textwidth}{!}{
    \begin{tabular}{ll | cccccc | c}
        \toprule
        Pretrain & Arch & C18 & C24 & C30 & 2C17 & 4C17 & 22C16 & Average \\
        \midrule
        Scratch & ResNet-18 & 3.7 $\pm$ 0.2 & 7.8 $\pm$ 0.3 & 30.6 $\pm$ 1.2 & 3.8 $\pm$ 0.2 & 4.9 $\pm$ 0.2 & 4.8 $\pm$ 0.2 & 9.3 $\pm$ 0.4 \\
        Scratch & Swin & 18.2 $\pm$ 0.7 & 29.1 $\pm$ 1.1 & 61.5 $\pm$ 1.9 & 17.7 $\pm$ 0.6 & 16.4 $\pm$ 0.6 & 17.2 $\pm$ 0.6 & 26.7 $\pm$ 0.9 \\
        Scratch & ViT & 18.5 $\pm$ 0.7 & 26.5 $\pm$ 1.1 & 56.7 $\pm$ 1.8 & 16.6 $\pm$ 0.7 & 18.0 $\pm$ 0.8 & 17.7 $\pm$ 0.8 & 25.7 $\pm$ 1.0 \\
        BYOL & ResNet-18 & 3.6 $\pm$ 0.2 & 7.8 $\pm$ 0.3 & 30.3 $\pm$ 1.2 & 3.7 $\pm$ 0.2 & 4.9 $\pm$ 0.2 & 4.8 $\pm$ 0.2 & 9.2 $\pm$ 0.4 \\
        DINO & ResNet-18 & 3.2 $\pm$ 0.2 & 7.3 $\pm$ 0.3 & 30.4 $\pm$ 1.3 & 3.6 $\pm$ 0.1 & 4.7 $\pm$ 0.2 & 4.8 $\pm$ 0.2 & 9.0 $\pm$ 0.4 \\
        MoCo & ResNet-18 & 3.1 $\pm$ 0.2 & 7.1 $\pm$ 0.3 & 29.4 $\pm$ 1.2 & 3.5 $\pm$ 0.2 & 4.7 $\pm$ 0.2 & 4.7 $\pm$ 0.2 & 8.8 $\pm$ 0.4 \\
        SimCLR & ResNet-18 & 3.3 $\pm$ 0.2 & 7.1 $\pm$ 0.3 & 29.3 $\pm$ 1.3 & 3.5 $\pm$ 0.2 & 4.8 $\pm$ 0.2 & 4.5 $\pm$ 0.2 & 8.8 $\pm$ 0.4 \\
        \bottomrule
    \end{tabular}
    }
    \caption{
    \textbf{Adsorption performance on full training sets for all molecules.}
    Extending the results in Table~\ref{tab:c18_full}, we show the fine-tuning results on all six molecules rather than just C18. 
    We find similar improvements from pretraining compared to fine-tuning across all molecule adsorption prediction tasks. 
    }
    \label{tab:app_full_train}
\end{table}

%% file: figs_tabs/tab_all_results.tex
\begin{table}[h]
    \centering
    \resizebox{\textwidth}{!}{
    \begin{tabular}{ll | cccccccc}
        \toprule
        Pretrain & Arch & 10k & 5k & 1k & 750 & 500 & 400 & 200 & 100 \\
        \midrule
        \multicolumn{10}{c}{\textit{Linear Probing}} \\
        \midrule
        ImageNet & ViT-2D & 48.7 $\pm$ 1.1 & 52.7 $\pm$ 1.1 & 63.2 $\pm$ 1.3 & 65.4 $\pm$ 1.3 & 69.7 $\pm$ 1.4 & 72.7 $\pm$ 1.4 & 79.7 $\pm$ 1.5 & 87.1 $\pm$ 1.7 \\
        Hand & MLP & 56.9 $\pm$ 1.7 & 103.0 $\pm$ 1.9 & 132.3 $\pm$ 2.3 & 146.6 $\pm$ 2.6 & 154.2 $\pm$ 2.8 & 150.7 $\pm$ 2.8 & 166.7 $\pm$ 2.8 & 171.3 $\pm$ 2.9 \\
        3DINO & ViT & 60.8 $\pm$ 1.2 & 63.4 $\pm$ 1.3 & 78.8 $\pm$ 1.5 & 82.5 $\pm$ 1.6 & 87.4 $\pm$ 1.6 & 92.5 $\pm$ 1.7 & 110.0 $\pm$ 2.0 & 111.8 $\pm$ 2.0 \\
        BYOL & ResNet & 34.3 $\pm$ 0.8 & 35.4 $\pm$ 0.8 & 41.8 $\pm$ 0.9 & 44.1 $\pm$ 1.0 & 51.9 $\pm$ 1.0 & 58.5 $\pm$ 1.2 & 74.6 $\pm$ 1.4 & 94.4 $\pm$ 1.7 \\
        DINO & ResNet & 44.1 $\pm$ 1.1 & 45.0 $\pm$ 1.1 & 50.5 $\pm$ 1.2 & 51.6 $\pm$ 1.2 & 54.2 $\pm$ 1.3 & 55.5 $\pm$ 1.3 & 62.1 $\pm$ 1.4 & 73.8 $\pm$ 1.6 \\
        MoCo & ResNet & \textbf{27.4 $\pm$ 0.8} & \textbf{28.3 $\pm$ 0.8} & \textbf{32.4 $\pm$ 0.9} & \textbf{33.4 $\pm$ 0.9} & \textbf{36.7 $\pm$ 0.9} & \textbf{38.1 $\pm$ 1.0} & \textbf{43.6 $\pm$ 1.0} & \textbf{51.5 $\pm$ 1.2} \\
        SimCLR & ResNet & \textit{28.5 $\pm$ 0.8} & \textit{29.7 $\pm$ 0.8} & \textit{35.7 $\pm$ 0.9} & \textit{37.0 $\pm$ 0.9} & \textit{41.8 $\pm$ 1.0} & \textit{41.5 $\pm$ 1.0} & \textit{48.8 $\pm$ 1.1} & \textit{60.0 $\pm$ 1.3} \\
        \midrule
        \multicolumn{10}{c}{\textit{Fine-Tuning}} \\
        \midrule
        Scratch & ResNet & 13.5 $\pm$ 0.5 & 17.1 $\pm$ 0.5 & 28.4 $\pm$ 0.8 & 32.6 $\pm$ 0.9 & 37.0 $\pm$ 0.9 & 61.1 $\pm$ 1.4 & 62.5 $\pm$ 1.3 & \textit{84.6 $\pm$ 2.0} \\
        Scratch & Swin & 31.1 $\pm$ 1.1 & 34.5 $\pm$ 1.1 & 128.2 $\pm$ 2.2 & 75.8 $\pm$ 1.8 & 164.9 $\pm$ 2.5 & 105.3 $\pm$ 2.0 & 141.0 $\pm$ 1.9 & 161.1 $\pm$ 2.9 \\
        Scratch & ViT & 40.4 $\pm$ 1.3 & 50.8 $\pm$ 1.6 & 78.1 $\pm$ 1.9 & 78.2 $\pm$ 2.0 & 85.1 $\pm$ 2.0 & 87.9 $\pm$ 2.0 & 95.7 $\pm$ 2.0 & 101.0 $\pm$ 2.2 \\
        BYOL & ResNet & \textit{12.4 $\pm$ 0.5} & \textit{16.5 $\pm$ 0.5} & \textit{22.6 $\pm$ 0.7} & \textit{25.9 $\pm$ 0.8} & \textit{29.8 $\pm$ 0.8} & \textit{47.8 $\pm$ 1.1} & 55.7 $\pm$ 1.3 & 86.9 $\pm$ 2.0 \\
        DINO & ResNet & 13.4 $\pm$ 0.4 & 19.3 $\pm$ 0.6 & 31.3 $\pm$ 0.9 & 31.6 $\pm$ 0.9 & 39.3 $\pm$ 1.0 & 54.2 $\pm$ 1.2 & 64.2 $\pm$ 1.5 & 85.1 $\pm$ 1.8 \\
        MoCo & ResNet & 12.6 $\pm$ 0.4 & 17.7 $\pm$ 0.5 & 23.0 $\pm$ 0.7 & 28.8 $\pm$ 0.8 & 36.2 $\pm$ 1.0 & 50.0 $\pm$ 1.2 & \textit{53.8 $\pm$ 1.4} & 100.8 $\pm$ 2.3 \\
        SimCLR & ResNet & \textbf{11.8 $\pm$ 0.4} & \textbf{14.9 $\pm$ 0.5} & \textbf{19.6 $\pm$ 0.6} & \textbf{20.4 $\pm$ 0.6} & \textbf{24.4 $\pm$ 0.7} & \textbf{37.1 $\pm$ 1.0} & \textbf{39.6 $\pm$ 1.1} & \textbf{61.8 $\pm$ 1.6} \\
        \bottomrule
    \end{tabular}
    }
    \caption{\textbf{Adsorption Performance ($\log(k_H)$ MSE, lower is better) across subsets for C18.} Best results are given in \textbf{bold} and second-best are given in \textit{italics}.}
    \label{tab:adsorption_subsets_c18}
\end{table}

\begin{table}[h]
    \centering
    \resizebox{\textwidth}{!}{
    \begin{tabular}{ll | cccccccc}
        \toprule
        Pretrain & Arch & 10k & 5k & 1k & 750 & 500 & 400 & 200 & 100 \\
        \midrule
        \multicolumn{10}{c}{\textit{Linear Probing}} \\
        \midrule
        ImageNet & ViT-2D & 48.7 $\pm$ 1.1 & 52.7 $\pm$ 1.1 & 63.2 $\pm$ 1.3 & 65.4 $\pm$ 1.3 & 69.7 $\pm$ 1.4 & 72.7 $\pm$ 1.4 & 79.7 $\pm$ 1.5 & 87.1 $\pm$ 1.7 \\
        Hand & MLP & 56.9 $\pm$ 1.7 & 103.0 $\pm$ 1.9 & 132.3 $\pm$ 2.3 & 146.6 $\pm$ 2.6 & 154.2 $\pm$ 2.8 & 150.7 $\pm$ 2.8 & 166.7 $\pm$ 2.8 & 171.3 $\pm$ 2.9 \\
        3DINO & ViT & 60.8 $\pm$ 1.2 & 63.4 $\pm$ 1.3 & 78.8 $\pm$ 1.5 & 82.5 $\pm$ 1.6 & 87.4 $\pm$ 1.6 & 92.5 $\pm$ 1.7 & 110.0 $\pm$ 2.0 & 111.8 $\pm$ 2.0 \\
        BYOL & ResNet & 34.3 $\pm$ 0.8 & 35.4 $\pm$ 0.8 & 41.8 $\pm$ 0.9 & 44.1 $\pm$ 1.0 & 51.9 $\pm$ 1.0 & 58.5 $\pm$ 1.2 & 74.6 $\pm$ 1.4 & 94.4 $\pm$ 1.7 \\
        DINO & ResNet & 44.1 $\pm$ 1.1 & 45.0 $\pm$ 1.1 & 50.5 $\pm$ 1.2 & 51.6 $\pm$ 1.2 & 54.2 $\pm$ 1.3 & 55.5 $\pm$ 1.3 & 62.1 $\pm$ 1.4 & 73.8 $\pm$ 1.6 \\
        MoCo & ResNet & \textbf{27.4 $\pm$ 0.8} & \textbf{28.3 $\pm$ 0.8} & \textbf{32.4 $\pm$ 0.9} & \textbf{33.4 $\pm$ 0.9} & \textbf{36.7 $\pm$ 0.9} & \textbf{38.1 $\pm$ 1.0} & \textbf{43.6 $\pm$ 1.0} & \textbf{51.5 $\pm$ 1.2} \\
        SimCLR & ResNet & \textit{28.5 $\pm$ 0.8} & \textit{29.7 $\pm$ 0.8} & \textit{35.7 $\pm$ 0.9} & \textit{37.0 $\pm$ 0.9} & \textit{41.8 $\pm$ 1.0} & \textit{41.5 $\pm$ 1.0} & \textit{48.8 $\pm$ 1.1} & \textit{60.0 $\pm$ 1.3} \\
        \midrule
        \multicolumn{10}{c}{\textit{Fine-Tuning}} \\
        \midrule
        Scratch & ResNet & 13.5 $\pm$ 0.5 & 17.1 $\pm$ 0.5 & 28.4 $\pm$ 0.8 & 32.6 $\pm$ 0.9 & 37.0 $\pm$ 0.9 & 61.1 $\pm$ 1.4 & 62.5 $\pm$ 1.3 & \textit{84.6 $\pm$ 2.0} \\
        Scratch & Swin & 43.1 $\pm$ 1.4 & 50.0 $\pm$ 1.7 & 162.2 $\pm$ 2.8 & 175.0 $\pm$ 2.7 & 196.2 $\pm$ 3.1 & 126.6 $\pm$ 2.6 & 178.1 $\pm$ 3.2 & 188.0 $\pm$ 2.8 \\
        Scratch & ViT & 40.4 $\pm$ 1.3 & 50.8 $\pm$ 1.6 & 78.1 $\pm$ 1.9 & 78.2 $\pm$ 2.0 & 85.1 $\pm$ 2.0 & 87.9 $\pm$ 2.0 & 95.7 $\pm$ 2.0 & 101.0 $\pm$ 2.2 \\
        BYOL & ResNet & \textit{12.4 $\pm$ 0.5} & \textit{16.5 $\pm$ 0.5} & \textit{22.6 $\pm$ 0.7} & \textit{25.9 $\pm$ 0.8} & \textit{29.8 $\pm$ 0.8} & \textit{47.8 $\pm$ 1.1} & 55.7 $\pm$ 1.3 & 86.9 $\pm$ 2.0 \\
        DINO & ResNet & 13.4 $\pm$ 0.4 & 19.3 $\pm$ 0.6 & 31.3 $\pm$ 0.9 & 31.6 $\pm$ 0.9 & 39.3 $\pm$ 1.0 & 54.2 $\pm$ 1.2 & 64.2 $\pm$ 1.5 & 85.1 $\pm$ 1.8 \\
        MoCo & ResNet & 12.6 $\pm$ 0.4 & 17.7 $\pm$ 0.5 & 23.0 $\pm$ 0.7 & 28.8 $\pm$ 0.8 & 36.2 $\pm$ 1.0 & 50.0 $\pm$ 1.2 & \textit{53.8 $\pm$ 1.4} & 100.8 $\pm$ 2.3 \\
        SimCLR & ResNet & \textbf{11.8 $\pm$ 0.4} & \textbf{14.9 $\pm$ 0.5} & \textbf{19.6 $\pm$ 0.6} & \textbf{20.4 $\pm$ 0.6} & \textbf{24.4 $\pm$ 0.7} & \textbf{37.1 $\pm$ 1.0} & \textbf{39.6 $\pm$ 1.1} & \textbf{61.8 $\pm$ 1.6} \\
        \bottomrule
    \end{tabular}
    }
    \caption{\textbf{Adsorption Performance ($\log(k_H)$ MSE, lower is better) across subsets for C24.} Best results are given in \textbf{bold} and second-best are given in \textit{italics}.}
    \label{tab:adsorption_subsets_c24}
\end{table}

\begin{table}[h]
    \centering
    \resizebox{\textwidth}{!}{
    \begin{tabular}{ll | cccccccc}
        \toprule
        Pretrain & Arch & 10k & 5k & 1k & 750 & 500 & 400 & 200 & 100 \\
        \midrule
        \multicolumn{10}{c}{\textit{Linear Probing}} \\
        \midrule
        ImageNet & ViT-2D & 48.7 $\pm$ 1.1 & 52.7 $\pm$ 1.1 & 63.2 $\pm$ 1.3 & 65.4 $\pm$ 1.3 & 69.7 $\pm$ 1.4 & 72.7 $\pm$ 1.4 & 79.7 $\pm$ 1.5 & 87.1 $\pm$ 1.7 \\
        Hand & MLP & 56.9 $\pm$ 1.7 & 103.0 $\pm$ 1.9 & 132.3 $\pm$ 2.3 & 146.6 $\pm$ 2.6 & 154.2 $\pm$ 2.8 & 150.7 $\pm$ 2.8 & 166.7 $\pm$ 2.8 & 171.3 $\pm$ 2.9 \\
        3DINO & ViT & 60.8 $\pm$ 1.2 & 63.4 $\pm$ 1.3 & 78.8 $\pm$ 1.5 & 82.5 $\pm$ 1.6 & 87.4 $\pm$ 1.6 & 92.5 $\pm$ 1.7 & 110.0 $\pm$ 2.0 & 111.8 $\pm$ 2.0 \\
        BYOL & ResNet & 34.3 $\pm$ 0.8 & 35.4 $\pm$ 0.8 & 41.8 $\pm$ 0.9 & 44.1 $\pm$ 1.0 & 51.9 $\pm$ 1.0 & 58.5 $\pm$ 1.2 & 74.6 $\pm$ 1.4 & 94.4 $\pm$ 1.7 \\
        DINO & ResNet & 44.1 $\pm$ 1.1 & 45.0 $\pm$ 1.1 & 50.5 $\pm$ 1.2 & 51.6 $\pm$ 1.2 & 54.2 $\pm$ 1.3 & 55.5 $\pm$ 1.3 & 62.1 $\pm$ 1.4 & 73.8 $\pm$ 1.6 \\
        MoCo & ResNet & \textbf{27.4 $\pm$ 0.8} & \textbf{28.3 $\pm$ 0.8} & \textbf{32.4 $\pm$ 0.9} & \textbf{33.4 $\pm$ 0.9} & \textbf{36.7 $\pm$ 0.9} & \textbf{38.1 $\pm$ 1.0} & \textbf{43.6 $\pm$ 1.0} & \textbf{51.5 $\pm$ 1.2} \\
        SimCLR & ResNet & \textit{28.5 $\pm$ 0.8} & \textit{29.7 $\pm$ 0.8} & \textit{35.7 $\pm$ 0.9} & \textit{37.0 $\pm$ 0.9} & \textit{41.8 $\pm$ 1.0} & \textit{41.5 $\pm$ 1.0} & \textit{48.8 $\pm$ 1.1} & \textit{60.0 $\pm$ 1.3} \\
        \midrule
        \multicolumn{10}{c}{\textit{Fine-Tuning}} \\
        \midrule
        Scratch & ResNet & 13.5 $\pm$ 0.5 & 17.1 $\pm$ 0.5 & 28.4 $\pm$ 0.8 & 32.6 $\pm$ 0.9 & 37.0 $\pm$ 0.9 & 61.1 $\pm$ 1.4 & 62.5 $\pm$ 1.3 & \textit{84.6 $\pm$ 2.0} \\
        Scratch & Swin & 76.4 $\pm$ 2.2 & 89.4 $\pm$ 2.4 & 126.3 $\pm$ 2.9 & 121.8 $\pm$ 2.8 & 124.3 $\pm$ 2.7 & 132.0 $\pm$ 2.7 & 163.4 $\pm$ 2.8 & 173.1 $\pm$ 2.9 \\
        Scratch & ViT & 40.4 $\pm$ 1.3 & 50.8 $\pm$ 1.6 & 78.1 $\pm$ 1.9 & 78.2 $\pm$ 2.0 & 85.1 $\pm$ 2.0 & 87.9 $\pm$ 2.0 & 95.7 $\pm$ 2.0 & 101.0 $\pm$ 2.2 \\
        BYOL & ResNet & \textit{12.4 $\pm$ 0.5} & \textit{16.5 $\pm$ 0.5} & \textit{22.6 $\pm$ 0.7} & \textit{25.9 $\pm$ 0.8} & \textit{29.8 $\pm$ 0.8} & \textit{47.8 $\pm$ 1.1} & 55.7 $\pm$ 1.3 & 86.9 $\pm$ 2.0 \\
        DINO & ResNet & 13.4 $\pm$ 0.4 & 19.3 $\pm$ 0.6 & 31.3 $\pm$ 0.9 & 31.6 $\pm$ 0.9 & 39.3 $\pm$ 1.0 & 54.2 $\pm$ 1.2 & 64.2 $\pm$ 1.5 & 85.1 $\pm$ 1.8 \\
        MoCo & ResNet & 12.6 $\pm$ 0.4 & 17.7 $\pm$ 0.5 & 23.0 $\pm$ 0.7 & 28.8 $\pm$ 0.8 & 36.2 $\pm$ 1.0 & 50.0 $\pm$ 1.2 & \textit{53.8 $\pm$ 1.4} & 100.8 $\pm$ 2.3 \\
        SimCLR & ResNet & \textbf{11.8 $\pm$ 0.4} & \textbf{14.9 $\pm$ 0.5} & \textbf{19.6 $\pm$ 0.6} & \textbf{20.4 $\pm$ 0.6} & \textbf{24.4 $\pm$ 0.7} & \textbf{37.1 $\pm$ 1.0} & \textbf{39.6 $\pm$ 1.1} & \textbf{61.8 $\pm$ 1.6} \\
        \bottomrule
    \end{tabular}
    }
    \caption{\textbf{Adsorption Performance ($\log(k_H)$ MSE, lower is better) across subsets for C30.} Best results are given in \textbf{bold} and second-best are given in \textit{italics}.}
    \label{tab:adsorption_subsets_c30}
\end{table}

\begin{table}[h]
    \centering
    \resizebox{\textwidth}{!}{
    \begin{tabular}{ll | cccccccc}
        \toprule
        Pretrain & Arch & 10k & 5k & 1k & 750 & 500 & 400 & 200 & 100 \\
        \midrule
        \multicolumn{10}{c}{\textit{Linear Probing}} \\
        \midrule
        ImageNet & ViT-2D & 48.7 $\pm$ 1.1 & 52.7 $\pm$ 1.1 & 63.2 $\pm$ 1.3 & 65.4 $\pm$ 1.3 & 69.7 $\pm$ 1.4 & 72.7 $\pm$ 1.4 & 79.7 $\pm$ 1.5 & 87.1 $\pm$ 1.7 \\
        Hand & MLP & 56.9 $\pm$ 1.7 & 103.0 $\pm$ 1.9 & 132.3 $\pm$ 2.3 & 146.6 $\pm$ 2.6 & 154.2 $\pm$ 2.8 & 150.7 $\pm$ 2.8 & 166.7 $\pm$ 2.8 & 171.3 $\pm$ 2.9 \\
        3DINO & ViT & 60.8 $\pm$ 1.2 & 63.4 $\pm$ 1.3 & 78.8 $\pm$ 1.5 & 82.5 $\pm$ 1.6 & 87.4 $\pm$ 1.6 & 92.5 $\pm$ 1.7 & 110.0 $\pm$ 2.0 & 111.8 $\pm$ 2.0 \\
        BYOL & ResNet & 34.3 $\pm$ 0.8 & 35.4 $\pm$ 0.8 & 41.8 $\pm$ 0.9 & 44.1 $\pm$ 1.0 & 51.9 $\pm$ 1.0 & 58.5 $\pm$ 1.2 & 74.6 $\pm$ 1.4 & 94.4 $\pm$ 1.7 \\
        DINO & ResNet & 44.1 $\pm$ 1.1 & 45.0 $\pm$ 1.1 & 50.5 $\pm$ 1.2 & 51.6 $\pm$ 1.2 & 54.2 $\pm$ 1.3 & 55.5 $\pm$ 1.3 & 62.1 $\pm$ 1.4 & 73.8 $\pm$ 1.6 \\
        MoCo & ResNet & \textbf{27.4 $\pm$ 0.8} & \textbf{28.3 $\pm$ 0.8} & \textbf{32.4 $\pm$ 0.9} & \textbf{33.4 $\pm$ 0.9} & \textbf{36.7 $\pm$ 0.9} & \textbf{38.1 $\pm$ 1.0} & \textbf{43.6 $\pm$ 1.0} & \textbf{51.5 $\pm$ 1.2} \\
        SimCLR & ResNet & \textit{28.5 $\pm$ 0.8} & \textit{29.7 $\pm$ 0.8} & \textit{35.7 $\pm$ 0.9} & \textit{37.0 $\pm$ 0.9} & \textit{41.8 $\pm$ 1.0} & \textit{41.5 $\pm$ 1.0} & \textit{48.8 $\pm$ 1.1} & \textit{60.0 $\pm$ 1.3} \\
        \midrule
        \multicolumn{10}{c}{\textit{Fine-Tuning}} \\
        \midrule
        Scratch & ResNet & 13.5 $\pm$ 0.5 & 17.1 $\pm$ 0.5 & 28.4 $\pm$ 0.8 & 32.6 $\pm$ 0.9 & 37.0 $\pm$ 0.9 & 61.1 $\pm$ 1.4 & 62.5 $\pm$ 1.3 & \textit{84.6 $\pm$ 2.0} \\
        Scratch & Swin & 27.7 $\pm$ 0.9 & 34.4 $\pm$ 1.1 & 60.1 $\pm$ 1.5 & 155.4 $\pm$ 2.4 & 89.4 $\pm$ 1.8 & 85.1 $\pm$ 1.8 & 98.3 $\pm$ 1.8 & 125.3 $\pm$ 2.1 \\
        Scratch & ViT & 40.4 $\pm$ 1.3 & 50.8 $\pm$ 1.6 & 78.1 $\pm$ 1.9 & 78.2 $\pm$ 2.0 & 85.1 $\pm$ 2.0 & 87.9 $\pm$ 2.0 & 95.7 $\pm$ 2.0 & 101.0 $\pm$ 2.2 \\
        BYOL & ResNet & \textit{12.4 $\pm$ 0.5} & \textit{16.5 $\pm$ 0.5} & \textit{22.6 $\pm$ 0.7} & \textit{25.9 $\pm$ 0.8} & \textit{29.8 $\pm$ 0.8} & \textit{47.8 $\pm$ 1.1} & 55.7 $\pm$ 1.3 & 86.9 $\pm$ 2.0 \\
        DINO & ResNet & 13.4 $\pm$ 0.4 & 19.3 $\pm$ 0.6 & 31.3 $\pm$ 0.9 & 31.6 $\pm$ 0.9 & 39.3 $\pm$ 1.0 & 54.2 $\pm$ 1.2 & 64.2 $\pm$ 1.5 & 85.1 $\pm$ 1.8 \\
        MoCo & ResNet & 12.6 $\pm$ 0.4 & 17.7 $\pm$ 0.5 & 23.0 $\pm$ 0.7 & 28.8 $\pm$ 0.8 & 36.2 $\pm$ 1.0 & 50.0 $\pm$ 1.2 & \textit{53.8 $\pm$ 1.4} & 100.8 $\pm$ 2.3 \\
        SimCLR & ResNet & \textbf{11.8 $\pm$ 0.4} & \textbf{14.9 $\pm$ 0.5} & \textbf{19.6 $\pm$ 0.6} & \textbf{20.4 $\pm$ 0.6} & \textbf{24.4 $\pm$ 0.7} & \textbf{37.1 $\pm$ 1.0} & \textbf{39.6 $\pm$ 1.1} & \textbf{61.8 $\pm$ 1.6} \\
        \bottomrule
    \end{tabular}
    }
    \caption{\textbf{Adsorption Performance ($\log(k_H)$ MSE, lower is better) across subsets for 2C17.} Best results are given in \textbf{bold} and second-best are given in \textit{italics}.}
    \label{tab:adsorption_subsets_2c17}
\end{table}

\begin{table}[h]
    \centering
    \resizebox{\textwidth}{!}{
    \begin{tabular}{ll | cccccccc}
        \toprule
        Pretrain & Arch & 10k & 5k & 1k & 750 & 500 & 400 & 200 & 100 \\
        \midrule
        \multicolumn{10}{c}{\textit{Linear Probing}} \\
        \midrule
        ImageNet & ViT-2D & 48.7 $\pm$ 1.1 & 52.7 $\pm$ 1.1 & 63.2 $\pm$ 1.3 & 65.4 $\pm$ 1.3 & 69.7 $\pm$ 1.4 & 72.7 $\pm$ 1.4 & 79.7 $\pm$ 1.5 & 87.1 $\pm$ 1.7 \\
        Hand & MLP & 56.9 $\pm$ 1.7 & 103.0 $\pm$ 1.9 & 132.3 $\pm$ 2.3 & 146.6 $\pm$ 2.6 & 154.2 $\pm$ 2.8 & 150.7 $\pm$ 2.8 & 166.7 $\pm$ 2.8 & 171.3 $\pm$ 2.9 \\
        3DINO & ViT & 60.8 $\pm$ 1.2 & 63.4 $\pm$ 1.3 & 78.8 $\pm$ 1.5 & 82.5 $\pm$ 1.6 & 87.4 $\pm$ 1.6 & 92.5 $\pm$ 1.7 & 110.0 $\pm$ 2.0 & 111.8 $\pm$ 2.0 \\
        BYOL & ResNet & 34.3 $\pm$ 0.8 & 35.4 $\pm$ 0.8 & 41.8 $\pm$ 0.9 & 44.1 $\pm$ 1.0 & 51.9 $\pm$ 1.0 & 58.5 $\pm$ 1.2 & 74.6 $\pm$ 1.4 & 94.4 $\pm$ 1.7 \\
        DINO & ResNet & 44.1 $\pm$ 1.1 & 45.0 $\pm$ 1.1 & 50.5 $\pm$ 1.2 & 51.6 $\pm$ 1.2 & 54.2 $\pm$ 1.3 & 55.5 $\pm$ 1.3 & 62.1 $\pm$ 1.4 & 73.8 $\pm$ 1.6 \\
        MoCo & ResNet & \textbf{27.4 $\pm$ 0.8} & \textbf{28.3 $\pm$ 0.8} & \textbf{32.4 $\pm$ 0.9} & \textbf{33.4 $\pm$ 0.9} & \textbf{36.7 $\pm$ 0.9} & \textbf{38.1 $\pm$ 1.0} & \textbf{43.6 $\pm$ 1.0} & \textbf{51.5 $\pm$ 1.2} \\
        SimCLR & ResNet & \textit{28.5 $\pm$ 0.8} & \textit{29.7 $\pm$ 0.8} & \textit{35.7 $\pm$ 0.9} & \textit{37.0 $\pm$ 0.9} & \textit{41.8 $\pm$ 1.0} & \textit{41.5 $\pm$ 1.0} & \textit{48.8 $\pm$ 1.1} & \textit{60.0 $\pm$ 1.3} \\
        \midrule
        \multicolumn{10}{c}{\textit{Fine-Tuning}} \\
        \midrule
        Scratch & ResNet & 13.5 $\pm$ 0.5 & 17.1 $\pm$ 0.5 & 28.4 $\pm$ 0.8 & 32.6 $\pm$ 0.9 & 37.0 $\pm$ 0.9 & 61.1 $\pm$ 1.4 & 62.5 $\pm$ 1.3 & \textit{84.6 $\pm$ 2.0} \\
        Scratch & Swin & 26.6 $\pm$ 0.9 & 36.8 $\pm$ 1.1 & 60.7 $\pm$ 1.5 & 85.8 $\pm$ 1.6 & 61.4 $\pm$ 1.4 & 128.2 $\pm$ 1.8 & 114.3 $\pm$ 2.0 & 94.5 $\pm$ 1.8 \\
        Scratch & ViT & 40.4 $\pm$ 1.3 & 50.8 $\pm$ 1.6 & 78.1 $\pm$ 1.9 & 78.2 $\pm$ 2.0 & 85.1 $\pm$ 2.0 & 87.9 $\pm$ 2.0 & 95.7 $\pm$ 2.0 & 101.0 $\pm$ 2.2 \\
        BYOL & ResNet & \textit{12.4 $\pm$ 0.5} & \textit{16.5 $\pm$ 0.5} & \textit{22.6 $\pm$ 0.7} & \textit{25.9 $\pm$ 0.8} & \textit{29.8 $\pm$ 0.8} & \textit{47.8 $\pm$ 1.1} & 55.7 $\pm$ 1.3 & 86.9 $\pm$ 2.0 \\
        DINO & ResNet & 13.4 $\pm$ 0.4 & 19.3 $\pm$ 0.6 & 31.3 $\pm$ 0.9 & 31.6 $\pm$ 0.9 & 39.3 $\pm$ 1.0 & 54.2 $\pm$ 1.2 & 64.2 $\pm$ 1.5 & 85.1 $\pm$ 1.8 \\
        MoCo & ResNet & 12.6 $\pm$ 0.4 & 17.7 $\pm$ 0.5 & 23.0 $\pm$ 0.7 & 28.8 $\pm$ 0.8 & 36.2 $\pm$ 1.0 & 50.0 $\pm$ 1.2 & \textit{53.8 $\pm$ 1.4} & 100.8 $\pm$ 2.3 \\
        SimCLR & ResNet & \textbf{11.8 $\pm$ 0.4} & \textbf{14.9 $\pm$ 0.5} & \textbf{19.6 $\pm$ 0.6} & \textbf{20.4 $\pm$ 0.6} & \textbf{24.4 $\pm$ 0.7} & \textbf{37.1 $\pm$ 1.0} & \textbf{39.6 $\pm$ 1.1} & \textbf{61.8 $\pm$ 1.6} \\
        \bottomrule
    \end{tabular}
    }
    \caption{\textbf{Adsorption Performance ($\log(k_H)$ MSE, lower is better) across subsets for 4C17.} Best results are given in \textbf{bold} and second-best are given in \textit{italics}.}
    \label{tab:adsorption_subsets_4c17}
\end{table}

\begin{table}[h]
    \centering
    \resizebox{\textwidth}{!}{
    \begin{tabular}{ll | cccccccc}
        \toprule
        Pretrain & Arch & 10k & 5k & 1k & 750 & 500 & 400 & 200 & 100 \\
        \midrule
        \multicolumn{10}{c}{\textit{Linear Probing}} \\
        \midrule
        ImageNet & ViT-2D & 48.7 $\pm$ 1.1 & 52.7 $\pm$ 1.1 & 63.2 $\pm$ 1.3 & 65.4 $\pm$ 1.3 & 69.7 $\pm$ 1.4 & 72.7 $\pm$ 1.4 & 79.7 $\pm$ 1.5 & 87.1 $\pm$ 1.7 \\
        Hand & MLP & 56.9 $\pm$ 1.7 & 103.0 $\pm$ 1.9 & 132.3 $\pm$ 2.3 & 146.6 $\pm$ 2.6 & 154.2 $\pm$ 2.8 & 150.7 $\pm$ 2.8 & 166.7 $\pm$ 2.8 & 171.3 $\pm$ 2.9 \\
        3DINO & ViT & 60.8 $\pm$ 1.2 & 63.4 $\pm$ 1.3 & 78.8 $\pm$ 1.5 & 82.5 $\pm$ 1.6 & 87.4 $\pm$ 1.6 & 92.5 $\pm$ 1.7 & 110.0 $\pm$ 2.0 & 111.8 $\pm$ 2.0 \\
        BYOL & ResNet & 34.3 $\pm$ 0.8 & 35.4 $\pm$ 0.8 & 41.8 $\pm$ 0.9 & 44.1 $\pm$ 1.0 & 51.9 $\pm$ 1.0 & 58.5 $\pm$ 1.2 & 74.6 $\pm$ 1.4 & 94.4 $\pm$ 1.7 \\
        DINO & ResNet & 44.1 $\pm$ 1.1 & 45.0 $\pm$ 1.1 & 50.5 $\pm$ 1.2 & 51.6 $\pm$ 1.2 & 54.2 $\pm$ 1.3 & 55.5 $\pm$ 1.3 & 62.1 $\pm$ 1.4 & 73.8 $\pm$ 1.6 \\
        MoCo & ResNet & \textbf{27.4 $\pm$ 0.8} & \textbf{28.3 $\pm$ 0.8} & \textbf{32.4 $\pm$ 0.9} & \textbf{33.4 $\pm$ 0.9} & \textbf{36.7 $\pm$ 0.9} & \textbf{38.1 $\pm$ 1.0} & \textbf{43.6 $\pm$ 1.0} & \textbf{51.5 $\pm$ 1.2} \\
        SimCLR & ResNet & \textit{28.5 $\pm$ 0.8} & \textit{29.7 $\pm$ 0.8} & \textit{35.7 $\pm$ 0.9} & \textit{37.0 $\pm$ 0.9} & \textit{41.8 $\pm$ 1.0} & \textit{41.5 $\pm$ 1.0} & \textit{48.8 $\pm$ 1.1} & \textit{60.0 $\pm$ 1.3} \\
        \midrule
        \multicolumn{10}{c}{\textit{Fine-Tuning}} \\
        \midrule
        Scratch & ResNet & 13.5 $\pm$ 0.5 & 17.1 $\pm$ 0.5 & 28.4 $\pm$ 0.8 & 32.6 $\pm$ 0.9 & 37.0 $\pm$ 0.9 & 61.1 $\pm$ 1.4 & 62.5 $\pm$ 1.3 & \textit{84.6 $\pm$ 2.0} \\
        Scratch & Swin & 25.5 $\pm$ 0.9 & 30.2 $\pm$ 1.0 & 154.7 $\pm$ 2.7 & 124.8 $\pm$ 2.6 & 65.3 $\pm$ 1.6 & 156.3 $\pm$ 3.0 & 87.0 $\pm$ 1.8 & 93.6 $\pm$ 1.8 \\
        Scratch & ViT & 40.4 $\pm$ 1.3 & 50.8 $\pm$ 1.6 & 78.1 $\pm$ 1.9 & 78.2 $\pm$ 2.0 & 85.1 $\pm$ 2.0 & 87.9 $\pm$ 2.0 & 95.7 $\pm$ 2.0 & 101.0 $\pm$ 2.2 \\
        BYOL & ResNet & \textit{12.4 $\pm$ 0.5} & \textit{16.5 $\pm$ 0.5} & \textit{22.6 $\pm$ 0.7} & \textit{25.9 $\pm$ 0.8} & \textit{29.8 $\pm$ 0.8} & \textit{47.8 $\pm$ 1.1} & 55.7 $\pm$ 1.3 & 86.9 $\pm$ 2.0 \\
        DINO & ResNet & 13.4 $\pm$ 0.4 & 19.3 $\pm$ 0.6 & 31.3 $\pm$ 0.9 & 31.6 $\pm$ 0.9 & 39.3 $\pm$ 1.0 & 54.2 $\pm$ 1.2 & 64.2 $\pm$ 1.5 & 85.1 $\pm$ 1.8 \\
        MoCo & ResNet & 12.6 $\pm$ 0.4 & 17.7 $\pm$ 0.5 & 23.0 $\pm$ 0.7 & 28.8 $\pm$ 0.8 & 36.2 $\pm$ 1.0 & 50.0 $\pm$ 1.2 & \textit{53.8 $\pm$ 1.4} & 100.8 $\pm$ 2.3 \\
        SimCLR & ResNet & \textbf{11.8 $\pm$ 0.4} & \textbf{14.9 $\pm$ 0.5} & \textbf{19.6 $\pm$ 0.6} & \textbf{20.4 $\pm$ 0.6} & \textbf{24.4 $\pm$ 0.7} & \textbf{37.1 $\pm$ 1.0} & \textbf{39.6 $\pm$ 1.1} & \textbf{61.8 $\pm$ 1.6} \\
        \bottomrule
    \end{tabular}
    }
    \caption{\textbf{Adsorption Performance ($\log(k_H)$ MSE, lower is better) across subsets for 22C16.} Best results are given in \textbf{bold} and second-best are given in \textit{italics}.}
    \label{tab:adsorption_subsets_22c16}
\end{table}

\begin{table}[h]
    \centering
    \resizebox{\textwidth}{!}{
    \begin{tabular}{ll | cccccccc}
        \toprule
        Pretrain & Arch & 10k & 5k & 1k & 750 & 500 & 400 & 200 & 100 \\
        \midrule
        \multicolumn{10}{c}{\textit{Linear Probing}} \\
        \midrule
        ImageNet & ViT-2D & 48.7 $\pm$ 1.1 & 52.7 $\pm$ 1.1 & 63.2 $\pm$ 1.3 & 65.4 $\pm$ 1.3 & 69.7 $\pm$ 1.4 & 72.7 $\pm$ 1.4 & 79.7 $\pm$ 1.5 & 87.1 $\pm$ 1.7 \\
        Hand & MLP & 56.9 $\pm$ 1.7 & 103.0 $\pm$ 1.9 & 132.3 $\pm$ 2.3 & 146.6 $\pm$ 2.6 & 154.2 $\pm$ 2.8 & 150.7 $\pm$ 2.8 & 166.7 $\pm$ 2.8 & 171.3 $\pm$ 2.9 \\
        3DINO & ViT & 60.8 $\pm$ 1.2 & 63.4 $\pm$ 1.3 & 78.8 $\pm$ 1.5 & 82.5 $\pm$ 1.6 & 87.4 $\pm$ 1.6 & 92.5 $\pm$ 1.7 & 110.0 $\pm$ 2.0 & 111.8 $\pm$ 2.0 \\
        BYOL & ResNet & 34.3 $\pm$ 0.8 & 35.4 $\pm$ 0.8 & 41.8 $\pm$ 0.9 & 44.1 $\pm$ 1.0 & 51.9 $\pm$ 1.0 & 58.5 $\pm$ 1.2 & 74.6 $\pm$ 1.4 & 94.4 $\pm$ 1.7 \\
        DINO & ResNet & 44.1 $\pm$ 1.1 & 45.0 $\pm$ 1.1 & 50.5 $\pm$ 1.2 & 51.6 $\pm$ 1.2 & 54.2 $\pm$ 1.3 & 55.5 $\pm$ 1.3 & 62.1 $\pm$ 1.4 & 73.8 $\pm$ 1.6 \\
        MoCo & ResNet & \textbf{27.4 $\pm$ 0.8} & \textbf{28.3 $\pm$ 0.8} & \textbf{32.4 $\pm$ 0.9} & \textbf{33.4 $\pm$ 0.9} & \textbf{36.7 $\pm$ 0.9} & \textbf{38.1 $\pm$ 1.0} & \textbf{43.6 $\pm$ 1.0} & \textbf{51.5 $\pm$ 1.2} \\
        SimCLR & ResNet & \textit{28.5 $\pm$ 0.8} & \textit{29.7 $\pm$ 0.8} & \textit{35.7 $\pm$ 0.9} & \textit{37.0 $\pm$ 0.9} & \textit{41.8 $\pm$ 1.0} & \textit{41.5 $\pm$ 1.0} & \textit{48.8 $\pm$ 1.1} & \textit{60.0 $\pm$ 1.3} \\
        \midrule
        \multicolumn{10}{c}{\textit{Fine-Tuning}} \\
        \midrule
        Scratch & ResNet & 13.5 $\pm$ 0.5 & 17.1 $\pm$ 0.5 & 28.4 $\pm$ 0.8 & 32.6 $\pm$ 0.9 & 37.0 $\pm$ 0.9 & 61.1 $\pm$ 1.4 & 62.5 $\pm$ 1.3 & \textit{84.6 $\pm$ 2.0} \\
        Scratch & Swin & 38.4 $\pm$ 1.2 & 45.9 $\pm$ 1.4 & 115.4 $\pm$ 2.3 & 123.1 $\pm$ 2.3 & 116.9 $\pm$ 2.2 & 122.3 $\pm$ 2.3 & 130.3 $\pm$ 2.2 & 139.3 $\pm$ 2.4 \\
        Scratch & ViT & 40.4 $\pm$ 1.3 & 50.8 $\pm$ 1.6 & 78.1 $\pm$ 1.9 & 78.2 $\pm$ 2.0 & 85.1 $\pm$ 2.0 & 87.9 $\pm$ 2.0 & 95.7 $\pm$ 2.0 & 101.0 $\pm$ 2.2 \\
        BYOL & ResNet & \textit{12.4 $\pm$ 0.5} & \textit{16.5 $\pm$ 0.5} & \textit{22.6 $\pm$ 0.7} & \textit{25.9 $\pm$ 0.8} & \textit{29.8 $\pm$ 0.8} & \textit{47.8 $\pm$ 1.1} & 55.7 $\pm$ 1.3 & 86.9 $\pm$ 2.0 \\
        DINO & ResNet & 13.4 $\pm$ 0.4 & 19.3 $\pm$ 0.6 & 31.3 $\pm$ 0.9 & 31.6 $\pm$ 0.9 & 39.3 $\pm$ 1.0 & 54.2 $\pm$ 1.2 & 64.2 $\pm$ 1.5 & 85.1 $\pm$ 1.8 \\
        MoCo & ResNet & 12.6 $\pm$ 0.4 & 17.7 $\pm$ 0.5 & 23.0 $\pm$ 0.7 & 28.8 $\pm$ 0.8 & 36.2 $\pm$ 1.0 & 50.0 $\pm$ 1.2 & \textit{53.8 $\pm$ 1.4} & 100.8 $\pm$ 2.3 \\
        SimCLR & ResNet & \textbf{11.8 $\pm$ 0.4} & \textbf{14.9 $\pm$ 0.5} & \textbf{19.6 $\pm$ 0.6} & \textbf{20.4 $\pm$ 0.6} & \textbf{24.4 $\pm$ 0.7} & \textbf{37.1 $\pm$ 1.0} & \textbf{39.6 $\pm$ 1.1} & \textbf{61.8 $\pm$ 1.6} \\
        \bottomrule
    \end{tabular}
    }
    \caption{\textbf{Adsorption Performance ($\log(k_H)$ MSE, lower is better) across subsets averaged over all molecules.} Best results are given in \textbf{bold} and second-best are given in \textit{italics}.}
    \label{tab:adsorption_subsets_average}
\end{table}

%% file: checklist.tex
\clearpage
\section*{NeurIPS Paper Checklist}

\begin{enumerate}

\item {\bf Claims}
    \item[] Question: Do the main claims made in the abstract and introduction accurately reflect the paper's contributions and scope?
    \item[] Answer: \answerYes{} %
    \item[] Justification: We outline a list of contributions in the abstract and introduction. \datasetname is outlined in Sections~\ref{sec:dataset} and~\ref{sec:app_dataset}. We include results which are presented in Sections~\ref{sec:results} and~\ref{sec:app_full_results} and justify our conclusions using these results. 

\item {\bf Limitations}
    \item[] Question: Does the paper discuss the limitations of the work performed by the authors?
    \item[] Answer: \answerYes{} %
    \item[] Justification: In Section~\ref{sec:limitations} we include a discussion of limitations and future work. 

\item {\bf Theory assumptions and proofs}
    \item[] Question: For each theoretical result, does the paper provide the full set of assumptions and a complete (and correct) proof?
    \item[] Answer: \answerNA{} %
    \item[] Justification: The paper does not include theoretical results.

    \item {\bf Experimental result reproducibility}
    \item[] Question: Does the paper fully disclose all the information needed to reproduce the main experimental results of the paper to the extent that it affects the main claims and/or conclusions of the paper (regardless of whether the code and data are provided or not)?
    \item[] Answer: \answerYes{} %
    \item[] Justification: The code and hyperparameter configurations are provided and the dataset is publicly available. 

\item {\bf Open access to data and code}
    \item[] Question: Does the paper provide open access to the data and code, with sufficient instructions to faithfully reproduce the main experimental results, as described in supplemental material?
    \item[] Answer: \answerYes{} %
    \item[] Justification: The data is publicly accessible and the code is provided. The code also includes all of the configurations which were used to generate the results. 

\item {\bf Experimental setting/details}
    \item[] Question: Does the paper specify all the training and test details (e.g., data splits, hyperparameters, how they were chosen, type of optimizer) necessary to understand the results?
    \item[] Answer: \answerYes{} %
    \item[] Justification: The dataset split information is given in Section~\ref{sec:supp_split} and the hyperparameters are in Section~\ref{sec:supp_hyperparameters}. The code also includes the configurations used to generate each of the results. 

\item {\bf Experiment statistical significance}
    \item[] Question: Does the paper report error bars suitably and correctly defined or other appropriate information about the statistical significance of the experiments?
    \item[] Answer: \answerYes{} %
    \item[] Justification: We report error bars through bootstrapping on the test set as described in Section~\ref{sec:training_eval_protocol}.

\item {\bf Experiments compute resources}
    \item[] Question: For each experiment, does the paper provide sufficient information on the computer resources (type of compute workers, memory, time of execution) needed to reproduce the experiments?
    \item[] Answer: \answerYes{} %
    \item[] Justification: We describe the computational resources used to generate the results in Sections~\ref{sec:hyperparameters} and~\ref{sec:supp_hyperparameters}.
    
\item {\bf Code of ethics}
    \item[] Question: Does the research conducted in the paper conform, in every respect, with the NeurIPS Code of Ethics \url{https://neurips.cc/public/EthicsGuidelines}?
    \item[] Answer: \answerYes{} %
    \item[] Justification: We do not use human subjects or participants in this work. The dataset is already publicly available and still in use. We do not anticipate any harmful consequences from our work, given we focus on benchmarking SSL methods for zeolites. Zeolites promote efficiency in chemical industrial processes and can lower carbon emissions. 

\item {\bf Broader impacts}
    \item[] Question: Does the paper discuss both potential positive societal impacts and negative societal impacts of the work performed?
    \item[] Answer: \answerYes{} %
    \item[] Justification: We briefly discuss the potential effects of improvements on \datasetname in Sections~\ref{sec:limitations} and~\ref{sec:conclusion}. Gaining a better understanding of zeolites can potentially improve the efficiency of chemical industrial processes and lower carbon emissions.
    
\item {\bf Safeguards}
    \item[] Question: Does the paper describe safeguards that have been put in place for responsible release of data or models that have a high risk for misuse (e.g., pre-trained language models, image generators, or scraped datasets)?
    \item[] Answer: \answerNA{} %
    \item[] Justification: We are not releasing data or models with high risk of misuse.

\item {\bf Licenses for existing assets}
    \item[] Question: Are the creators or original owners of assets (e.g., code, data, models), used in the paper, properly credited and are the license and terms of use explicitly mentioned and properly respected?
    \item[] Answer: \answerYes{} %
    \item[] Justification: We explicitly cite works which we utilize and discuss in this paper. We credit the data processing and representation strategy to ZeoNet~\cite{liu2023zeonet} and appropriately cite models which we utilize.

\item {\bf New assets}
    \item[] Question: Are new assets introduced in the paper well documented and is the documentation provided alongside the assets?
    \item[] Answer: \answerYes{} %
    \item[] Justification: We release code which includes documentation in the README with information on how to replicate and extend our experimentation on \datasetname. 

\item {\bf Crowdsourcing and research with human subjects}
    \item[] Question: For crowdsourcing experiments and research with human subjects, does the paper include the full text of instructions given to participants and screenshots, if applicable, as well as details about compensation (if any)? 
    \item[] Answer: \answerNA{} %
    \item[] Justification: There are no crowdsourcing or human subjects in this work.

\item {\bf Institutional review board (IRB) approvals or equivalent for research with human subjects}
    \item[] Question: Does the paper describe potential risks incurred by study participants, whether such risks were disclosed to the subjects, and whether Institutional Review Board (IRB) approvals (or an equivalent approval/review based on the requirements of your country or institution) were obtained?
    \item[] Answer: \answerNA{} %
    \item[] Justification: There are no crowdsourcing or human subjects in this work.

\item {\bf Declaration of LLM usage}
    \item[] Question: Does the paper describe the usage of LLMs if it is an important, original, or non-standard component of the core methods in this research? Note that if the LLM is used only for writing, editing, or formatting purposes and does \emph{not} impact the core methodology, scientific rigor, or originality of the research, declaration is not required.
    \item[] Answer: \answerNA{} %
    \item[] Justification: The core method development does not involve LLMs as an important, original, or non-standard component. 

\end{enumerate}